\PassOptionsToPackage{unicode}{hyperref}
\PassOptionsToPackage{hyphens}{url}
\PassOptionsToPackage{dvipsnames,svgnames,x11names}{xcolor}
\documentclass[
  12pt]{article}

\usepackage{amsmath,amssymb,amsthm}
\usepackage{iftex}
\ifPDFTeX
  \usepackage[T1]{fontenc}
  \usepackage[utf8]{inputenc}
  \usepackage{textcomp} 
\else 
  \usepackage{unicode-math}
  \defaultfontfeatures{Scale=MatchLowercase}
  \defaultfontfeatures[\rmfamily]{Ligatures=TeX,Scale=1}
\fi
\usepackage{lmodern}
\ifPDFTeX\else  
\fi
\IfFileExists{upquote.sty}{\usepackage{upquote}}{}
\IfFileExists{microtype.sty}{
  \usepackage[]{microtype}
  \UseMicrotypeSet[protrusion]{basicmath} 
}{}
\makeatletter
\@ifundefined{KOMAClassName}{
  \IfFileExists{parskip.sty}{%
    \usepackage{parskip}
  }{
    \setlength{\parindent}{0pt}
    \setlength{\parskip}{6pt plus 2pt minus 1pt}}
}{
  \KOMAoptions{parskip=half}}
\makeatother
\usepackage{xcolor}
\makeatletter
\ifx\paragraph\undefined\else
  \let\oldparagraph\paragraph
  \renewcommand{\paragraph}{
    \@ifstar
      \xxxParagraphStar
      \xxxParagraphNoStar
  }
  \newcommand{\xxxParagraphStar}[1]{\oldparagraph*{#1}\mbox{}}
  \newcommand{\xxxParagraphNoStar}[1]{\oldparagraph{#1}\mbox{}}
\fi
\ifx\subparagraph\undefined\else
  \let\oldsubparagraph\subparagraph
  \renewcommand{\subparagraph}{
    \@ifstar
      \xxxSubParagraphStar
      \xxxSubParagraphNoStar
  }
  \newcommand{\xxxSubParagraphStar}[1]{\oldsubparagraph*{#1}\mbox{}}
  \newcommand{\xxxSubParagraphNoStar}[1]{\oldsubparagraph{#1}\mbox{}}
\fi
\makeatother

\usepackage{longtable,booktabs,array}
\usepackage{calc} 
\usepackage{etoolbox}
\makeatletter
\patchcmd\longtable{\par}{\if@noskipsec\mbox{}\fi\par}{}{}
\makeatother
\IfFileExists{footnotehyper.sty}{\usepackage{footnotehyper}}{\usepackage{footnote}}
\makesavenoteenv{longtable}
\usepackage{graphicx}
\makeatletter
\def\maxwidth{\ifdim\Gin@nat@width>\linewidth\linewidth\else\Gin@nat@width\fi}
\def\maxheight{\ifdim\Gin@nat@height>\textheight\textheight\else\Gin@nat@height\fi}
\makeatother
\setkeys{Gin}{width=\maxwidth,height=\maxheight,keepaspectratio}
\makeatletter
\def\fps@figure{htbp}
\makeatother

\makeatletter
\@ifpackageloaded{caption}{}{\usepackage{caption}}
\AtBeginDocument{%
\ifdefined\contentsname
  \renewcommand*\contentsname{Table of contents}
\else
  \newcommand\contentsname{Table of contents}
\fi
\ifdefined\listfigurename
  \renewcommand*\listfigurename{List of Figures}
\else
  \newcommand\listfigurename{List of Figures}
\fi
\ifdefined\listtablename
  \renewcommand*\listtablename{List of Tables}
\else
  \newcommand\listtablename{List of Tables}
\fi
\ifdefined\figurename
  \renewcommand*\figurename{Figure}
\else
  \newcommand\figurename{Figure}
\fi
\ifdefined\tablename
  \renewcommand*\tablename{Table}
\else
  \newcommand\tablename{Table}
\fi
}
\@ifpackageloaded{float}{}{\usepackage{float}}
\floatstyle{ruled}
\@ifundefined{c@chapter}{\newfloat{codelisting}{h}{lop}}{\newfloat{codelisting}{h}{lop}[chapter]}
\floatname{codelisting}{Listing}

\makeatother
\makeatletter
\@ifpackageloaded{caption}{}{\usepackage{caption}}
\@ifpackageloaded{subcaption}{}{\usepackage{subcaption}}
\makeatother

\ifLuaTeX
  \usepackage{selnolig}  
\fi
\usepackage[]{natbib}
\usepackage{bookmark}

\IfFileExists{xurl.sty}{\usepackage{xurl}}{} 
\hypersetup{
  pdftitle={Self-Supervised Learning with Gaussian Processes},
  pdfauthor={Author 1; Author 2},
  pdfkeywords={3 to 6 keywords, that do not appear in the title},
  colorlinks=true,
  linkcolor={blue},
  filecolor={Maroon},
  citecolor={Blue},
  urlcolor={Blue},
  pdfcreator={LaTeX via pandoc}}

\newcommand{\anon}{1}

\newtheorem{prop}{Proposition}

\newcommand{\eps}{\epsilon}

\newcommand{\invloss}{\ell^{\text{\small{inv}}}}
\newcommand{\covloss}{\ell^{\text{\small{cov}}}}
\newcommand{\varloss}{\ell^{\text{\small{var}}}}

\newcommand{\Zb}{\bm{Z}}

\usepackage{cleveref}

\crefname{prop}{proposition}{propositions}
\Crefname{prop}{Proposition}{Propositions}
\begin{document}

\def\spacingset#1{\renewcommand{\baselinestretch}%
{#1}\small\normalsize} \spacingset{1}


\if1\anon
{
  \title{\bf Self-Supervised Learning with Gaussian Processes}
  \author{Yunshan Duan
  \hspace{.2cm}\\
    Department of Applied Mathematics and Statistics, Johns Hopkins University\\
    and \\
    Sinead Williamson\\
    Apple}
  \maketitle
} \fi

\if0\anon
{
  \bigskip
  \bigskip
  \bigskip
  \begin{center}
    {\LARGE\bf Self-Supervised Learning with Gaussian Processes}
\end{center}
  \medskip
} \fi

\bigskip
\begin{abstract}
    Self supervised learning (SSL) is a machine learning paradigm where models learn to understand the underlying structure of data without explicit supervision from labeled samples. The acquired representations from SSL have demonstrated useful for many downstream tasks including clustering, and linear classification, etc. To ensure smoothness of the representation space, most SSL methods rely on the ability to generate pairs of observations that are similar to a given instance. However, generating these pairs may be challenging for many types of data. Moreover, these methods lack consideration of uncertainty quantification and can perform poorly in out-of-sample prediction settings. To address these limitations, we propose Gaussian process self supervised learning (GPSSL), a novel approach that utilizes Gaussian processes (GP) models on representation learning. GP priors are imposed on the representations, and we obtain a generalized Bayesian posterior minimizing a loss function that encourages informative representations. The covariance function inherent in GPs naturally pulls representations of similar units together, serving as an alternative to using explicitly defined positive samples. We show that GPSSL is closely related to both kernel PCA and VICReg, a popular neural network-based SSL method, but unlike both allows for posterior uncertainties that can be propagated to downstream tasks. Experiments on various datasets, considering classification and regression tasks, demonstrate that GPSSL outperforms traditional methods in terms of accuracy, uncertainty quantification, and error control.
\end{abstract}

\noindent%
\vfill

\newpage
\spacingset{1.8} 

\section{Introduction}

Self supervised learning (SSL) aims to learn representations by understanding the structure of data without supervision of labels. The representations learned by SSL have been shown to be useful for many downstream tasks including unsupervised clustering, linear classification, and object detection \citep{bachman2019learning, misra2020self, caron2020unsupervised, gidaris2021obow}. 
Contrastive approaches such as SimCLR \citep{chen2020simple} explicitly encode the idea that similar objects should have similar representations, and dissimilar objects should have dissimilar representations. This is enforced by identifying “positive pairs” of similar objects and “negative pairs” of dissimilar objects, and constructing a loss that encourages similarity between positive pairs while penalizing similarity between negative pairs. The positive pairs are generated by applying data augmentation techniques such as blurring, rotating or cropping images; the negative pairs are typically random pairings from the full dataset. A contrastive loss allows us to learn a joint embedding architecture where the representations for a sample and its positive pair are close to each other, while other negative samples are pushed away. 

Conversely, non-contrastive learning methods have been proposed where only positive samples are needed for training the representations, for example, BYOL \citep{grill2020bootstrap}, SimSiam \citep{chen2021exploring}, and VICReg \citep{bardes2021vicreg}. Instead of pushing negative samples away, they encourage representations to be different for distinct subjects by controlling the overall diversity of the representations, for example, through normalization, stop-gradient operations, or information maximization. Of particular relevance to our work, VICReg specifies a three-part loss: an invariance term that encourages positive pairs to have similar representations; a variance term that encourages diversity by rewarding high standard deviation of representations within a batch; and a covariance term that decorrelates the variables for each representation.
In general, these methods yield good performance on representation learning. However, while these approaches no longer require negative pairs, they still assume the availability of a data augmentation mechanism capable of generating positive pairs. Such data augmentation techniques have been well explored in image modeling, but are harder to define in other application areas such as time series analysis or tabular data. 

We avoid the need to specify positive pairs, by using a Gaussian process to enforce smoothness of the representations. Here, the smoothness is derived from an appropriate kernel defined on the space of observations; kernels can be specified for a wide range of objects, including graphs, texts and tabular data. We combine our Gaussian process prior with a loss comprising the variance and covariance losses from VICReg. This leads to a generalized Bayesian posterior, which we can (approximately) infer using generalized variational inference.

The result is a conditional distribution over representations, rather than the single representation obtained using standard SSL approaches. Incorporating uncertainty into our representations allows us to propagate this uncertainty into downstream tasks, which we show yields improved uncertainty estimation and better robustness. We show connections to kernel PCA, and in doing so show that SSL methods such as VICReg are highly related to kernel-based methods such as kernel PCA, complementing existing work on spectral interpretations of SSL \citep{balestriero2022contrastive,shwartz2023information}.

\subsection{Notation}
The goal of 
SSL is to learn a mapping $f_z:\mathcal{X}\rightarrow \mathbb{R}^J$ from some observational space $\mathcal{X}$ to a low-dimensional representation space. When working with probabilistic SSL methods, we will define a distribution over such functions, implying a predictive distribution over representations for each observation. We let $X = (x_1,\dots, x_N)$ denote our training set, and $Z = (z_1,\dots, z_N)$ denote the corresponding representations $z_i=f_z(x_i)$. We let $\bar{z} = \frac{1}{N} \sum_i z_i$ be the mean representation, and let $z^j$ be the column of $Z$ corresponding to the $j$th dimension. We use $x^*$ to denote an observation not in our training set, and $z^*=f_z(x^*)$ its corresponding representation.

We let $k(x_i, x_j)$ be a kernel function, and $K(X,X)$ be the associated Gram matrix evaluated on $X\times X$.

\subsection{Self-supervised learning}

Our method follows in the traditions of contrastive and non-contrastive SSL, whereby representational geometry is informed by knowledge about which observations are similar or dissimilar. Contrastive learning approaches, such as SimCLR \citep{chen2020simple} and MoCo \citep{he2020momentum}, aim to increase the representation similarity of similar observations (positive pairs), and decrease representational similarity of dissimilar observations (negative pairs). Positive pairs are typically obtained by applying a distortion to a reference observation. In an image setting, this might be via rotation or cropping \citep{chen2020simple}. In a tabular setting, we might add Gaussian noise \citep{bahri2021scarf} or replace random features with their mean or modal value \citep{ucar2021subtab}. Negative pairs are typically randomly sampled from the dataset. 

Non-contrastive methods, such as SimSiam \citep{chen2021exploring}, DINO \citep{caron2021emerging}, BYOL \citep{grill2020bootstrap}, and VICReg \citep{bardes2021vicreg}, also encourage representational similarity between positive pairs, but do not explicitly enforce representational dissimilarity between negative pairs. Instead, they introduce alternative mechanisms designed to avoid representational collapse, such as paired networks with stop gradients \citep{chen2020simple}. Our method is most closely related to VICReg \citep{bardes2021vicreg}, which minimizes a loss function $\ell = c_I\invloss + c_V\varloss + c_C\covloss$ with three terms: an \textbf{invariance loss}
\begin{equation}
    \invloss(Z, Z') = \frac{1}{N} ||z_i - z_i'||_2^2, \label{eqn:invloss}
\end{equation}
where $z_i'$ is the positive pair of $z_i$, that encourages representations to be similar when input data are close to each other; a \textbf{variance loss}
\begin{equation}\varloss(Z) = \frac{1}{J}\sum_{j=1}^J \max\left(0, \gamma-\sqrt{\text{Var}(z^j)+\epsilon}\right),\label{eqn:varloss}\end{equation}
that encourages a high standard deviation between representations, encouraging differentiation between representations; and a \textbf{covariance loss}
\begin{equation}\covloss(Z) = \frac{1}{N-1}\sum_{i=1}^N \left(z_i-\bar{z}\right)\left(z_i - \bar{z}\right)^T,\label{eqn:covloss}\end{equation}
which shrinks the covariance between each dimension of the representation, preventing information collapse due to correlated variables. \citet{shwartz2023information} discusses a connection between the VICReg objective and information-theoretic optimization, and emphasizing its practical advantages over other SSL methods.

Most work on SSL, including the methods discussed above, learns a deterministic mapping from observations to representations. A number of probabilistic variants have been proposed, that yield a distribution over representations which can capture representational uncertainty. Bootstrap Your Own Variance  \citep[BYOV,][]{bootstrap-own-variance} replaces the deterministic neural networks in BYOL with Bayesian neural networks.  VI-SimSiam \citep{nakamura2023representation} uses a variational inference-based approach to impose a distribution on the representational space. SimVAE \citep{bizeul2024probabilistic} posits a generative model underlying positive pairs, and uses variational inference to infer latent representations. Our work shares some similarity with these methods in that we obtain a distribution over observations; however unlike these approaches we do not rely on positive pairs to enforce similarity. Instead, we assume the existance of an appropriate kernel. This makes methods like BYOV, VI-SimSiam and SimVAE appropriate for high-dimensional data where a good augmentation model is available, such as images. Meanwhile, our approach is better suited to lower-dimensional representations where we do not have an obvious augmentation model but where similarities can be expressed via a kernel, such as tabular data and structured data such as text or networks.

\subsection{Gaussian Processes}
The Gaussian process (GP) is a convenient and powerful prior distribution on functions, $f: \mathcal{X} \rightarrow \mathbb{R}$. Specifically, a GP is a collection of random variables $\{f(x) \mid x \in \mathcal{X}\}$ for which, given any finite set of $N$ inputs $X = (x_1, \dots, x_N)$ the collection $Z = (f(x_1), \dots, f(x_N))$ has a joint multivariate Gaussian distribution. It is completely defined by its mean $m(\cdot)$ and covariance functions $k(\cdot, \cdot)$, and the elegant marginalization properties of the Gaussian distribution allow us to compute marginals and conditionals in closed form \citep[see e.g.,][]{williams2006gaussian}. The covariance function, or kernel,  enforces smoothness of samples from the Gaussian process and encodes assumptions about pairwise similarities of random variables. Covariance functions can be defined for a wide range of data types, including vectors in $\mathbb{R}^d$  \citep[see e.g.,][]{williams2006gaussian}, text \citep{lodhi2002text}, and graphs \citep{vishwanathan2010graph}. 


\subsection{PCA and Kernel PCA}

The SSL methods described above learn representations by enforcing data-driven similarity and dissimilarity constraints. An alternative approach is to construct representations that are informative about the original observations. For example, principal components analysis (PCA) linearly projects observations into a series of uncorrelated variables, or principal components. 

Kernel PCA extends this idea to incorporate non-linear projections \citep{scholkopf1998nonlinear}. Conceptually, kernel PCA projects observations to the reproducing kernel Hilbert space $\Phi$ defined by a given kernel, performs PCA in this space, and projects the principal components back to the original space. 

Other methods that fit this paradigm include variational autoencoders \citep{kingma2013auto} and nonnegative matrix factorization \citep{lee1999learning}.



\subsection{Generalized Bayes}

Classical Bayesian inference posits a prior distribution over latent parameters, and a likelihood model that describes how observations are generated given those parameters. Bayes' theorem then specifies the posterior distribution over parameters given the data. Generalized Bayesian inference replaces the likelihood model with an arbitrary loss function $\ell(x;\theta)$ \citep{bissiri2016general,jewson2018principles,knoblauch2019generalized}. We then obtain the generalized posterior as
$\tilde{p}(\theta|x) \propto p(\theta) \exp\{-\ell(x; \theta)\}.$


\section{A Gaussian process-based self-supervised learning algorithm}

In this section, we propose a new SSL method based on Gaussian processes, that avoids the need to specify either positive or negative pairs. Our approach, GPSSL, uses a Gaussian process to encourage representational similarity, and simple variance-based losses to avoid collapse. In \Cref{sec:GPSSL} we describe our method, and discuss how we obtain an approximate generalized posterior distribution over representations in \Cref{sec:inference}. In \Cref{sec:connection_to_ssl}, we discuss the relationship between GPSSL and VIGReg, and in \Cref{sec:kPCA}  we discuss the relationship between GPSSL and kernel PCA, additionally providing insights into the relationship between kernel PCA and other non-contrastive SSL methods.

\subsection{The GPSSL model}\label{sec:GPSSL}

Given unlabeled data $X=(x_1,\dots, x_N) \in \mathcal{X}^N$, and a kernel function $k: \mathcal{X}\times \mathcal{X}\rightarrow \mathbb{R}_+$, we wish to learn a (joint) distribution over representations $z^*\in \mathbb{R}^J$ for arbitrary $x^*\in \mathcal{X}$---or equivalently, a distribution over functions $f_z: \mathcal{X}\rightarrow \mathbb{R}^J$.

Following VICReg \citep{bardes2021vicreg}, we aim to learn representations that are smooth (c.f.\ VICReg's invariance loss); that vary across $\mathcal{X}$ (c.f.\ VICReg's variance loss); and which have decorrelated dimensions (c.f.\ VICReg's covariance loss). To ensure that the function $f_z$ is smoothly varying with respect to $x$, we place a Gaussian process prior on $f_z$, 
$$f_z(\cdot)\sim \mbox{GP}\left(m(\cdot), k(\cdot, \cdot)\right),$$
where we assume the mean function $m(x)=0$ without loss of generality.

While we have defined a prior on $f_z$, we do not have labels $Y$, so it is meaningless to define a traditional likelihood $p(Y|Z, X)$ and compute the standard Bayesian posterior. Instead, we consider a generalized posterior taking the form 
$\tilde{p} (f_z \mid X) \propto p(f_z) \exp \{ - \ell(Z:=f_z(X)) \}$.  
We use a VICReg-like objective function for $\ell$, 
\begin{equation}\ell(Z) = c_V\varloss(Z) + c_C\covloss(Z),\label{eqn:loss}\end{equation}
with $\varloss$ and $\covloss$ defined in \Cref{eqn:varloss,eqn:covloss} and with constants $c_V$ and $c_C$ determining their relative contributions. We exclude the invariance loss (\Cref{eqn:invloss}) since its contribution is captured by the GP priors, as we elaborate in \Cref{sec:connection_to_ssl}.

\subsection{Approximating the generalized Bayesian posterior} 
\label{sec:inference}
We approximate the generalized Bayesian posterior distribution $\tilde{p}$ using generalized variational inference \citep{knoblauch2019generalized}.
Similar to variational inference for the classical Bayesian posterior, we approximate the generalized Bayesian posterior $\tilde{p}(f_z) \propto p(f_z) \exp \{ - \ell(Z) \}$ by finding the closest distribution within a variational family $\mathcal{Q}$, such that
$$q^\star(f_z) = \arg \min_{q \in \mathcal{Q}} \left\{ E_{q(f_z)} \ell (Z) + \text{KL}(q || p) \right\},$$
where $\ell(\cdot)$ is the loss function described in \Cref{eqn:loss} and KL denotes the Kullback-Leibler divergence. 

Following \cite{hensman2015scalable}, we augment our model with a relatively small set of inducing points $(U_x, U_z)$, such that $[Z^T \; U_z^T]^T$ are samples from our Gaussian process evaluated at $[X^T \; U_x^T]^T$. We then specify our variational family via a Gaussian distribution over $U_z$. The generalized ELBO becomes
$$\text{ELBO} = -\mathbb{E}_{q(U_z)}[\ell(Z)] - \text{KL}\left(q(U_z) || p(U_z)\right) \geq -\mathbb{E}_{q(U_z)}\left[\mathbb{E}_{p(Z|U_Z)}\left[\ell(Z)\right]\right] - \text{KL}\left(q(U_z) || p(U_z)\right).$$
Note, this directly mirrors the standard ELBO in \cite{hensman2015scalable}, with our loss $\ell$ replacing the log likelihood. We estimate the expected loss using Monte Carlo samples.

\subsection{Relationship to other models}\label{sec:relationships}
In this section, we discuss the connection to two related approaches: VICReg (\Cref{sec:connection_to_ssl}) and Kernel PCA (\Cref{sec:kPCA}).

\subsubsection{The GP prior acts similarly to existing SSL invariance losses}\label{sec:connection_to_ssl}

As described in \Cref{sec:GPSSL}, GPSSL inherits the variance and covariance losses from VICReg (\Cref{eqn:varloss,eqn:covloss}), but does not include the invariance loss (\Cref{eqn:invloss}). In fact, unlike other contrastive and non-contrastive SSL methods, there is no explicit invariance loss that looks at the similarity between an observation and its positive pair.

This is possible because the Gaussian process prior provides a similar inductive bias to other SSL invariance losses. To see this, consider an augmented dataset $\tilde{X} = \left((x_1, x'_1), \dots, (x_N, x'_N)\right)$ consisting of our original data $X=(x_1,\dots, x_N)$, plus generated positive pairs $X'=(x'_1, \dots, x'_N)$.
If we let
\begin{equation}k(x_i, x_j)=\begin{cases} 1 & \text{ iff }x_j=x_i'\text{, i.e., if }x_i \text{ and }x_j\text{ form a positive pair}\\ 0 & \text{ otherwise,}\end{cases}\label{eqn:SSL_kernel}\end{equation}

then the log joint of $Z$ under the Gaussian process prior takes the form
$$\log p(Z) = \text{const} - \frac{1}{2}Z^TK^{-1}Z.$$
Substituting in \Cref{eqn:SSL_kernel}, this can be interpreted as a loss of the form $\sum_{i=1}^N z_i^Tz_i'$. This is a plausible SSL-like invariance loss; while we are not aware of any SSL methods that use an unnormalized dot product loss, a normalized dot product loss is a fairly common invariance loss in non-contrastive SSL, e.g.\ \citet{chen2021exploring}. 

From this, we see that the GP prior performs a similar role to the positive pairs used in VICReg and other SSL methods, and so we do not include positive pairs, relying instead on the GP prior to enforce local smoothness in the representations. We note that if we use the popular squared exponential kernel, samples from our GP will be infinitely differentiable. Alternatively, a data-driven kernel such as the nearest neighbor Gaussian process \citep{datta2016hierarchical,wu2022variational} would more explicitly capture the ``positive pairs''-type intuition that an observation's representation should be similar to that of its nearest neighbors.

\subsubsection{Connection to kernel PCA}\label{sec:kPCA}
As we have described above, GPSSL can be seen as a probabilistic, kernel-based variant of VICReg. In this section, we show that GPSSL can also be thought of as an (approximate) probabilistic variant of kernel PCA, providing a link between SSL methods such as VICReg and kernel-based dimensionality reduction methods.



First, consider conducting kernel PCA with a given kernel
$k(x_i,x_j) =\phi(x_i)^T\phi(x_j)$. Kernel PCA maps observations $x\in \mathcal{X}$ to a hidden embedding $\phi(x)\in \Phi$, performs PCA in the space $\Phi$, and projects the principal components down to $\mathcal{X}$. The kernelized principal component $z^*$ for test point $x^*$ is 
$z^* = \phi(x^*)^T v$ where $v$ is 
the solution to 
\begin{equation}\max_{v \in \Phi} \frac{1}N v^T \phi(X) \phi(X)^T v, \;\; s.t. \;\; v^T v = 1.\label{eqn:primal}\end{equation}
W.l.o.g., we assume the data is centered in the hidden space, $\sum_{i=1}^N \phi(x_i) = 0$, otherwise an augmentation of the kernel matrix is needed.

Now, consider using GPSSL to learn a one-dimensional representation, using the same kernel $k(\cdot, \cdot)$. If we replace the variance loss in \Cref{eqn:loss} with a similar but less robust alternative, \Cref{prop:kPCA} proves that there exists a value of $c_V$ for which the maximal point of the generalized posterior is equivalent to the first principal component of kernel PCA.


\begin{prop} Let the number of dimensions $J=1$, and replace the variance loss term in \Cref{eqn:loss} with  $V(Z) = - \mbox{Var}(Z) = \frac{1}N (Z - \overline{z})^T (Z - \overline{z})$. Then there exists a value of $c_V$ for which the generalized posterior is maximized at the first kernel PCA component. \label{prop:kPCA}
\end{prop}
\begin{proof}
We can write the dual problem of \Cref{eqn:primal} as 
\begin{equation} \label{eqn:kPCA}
    \max_{\lambda, \; v}  \frac{1}N v^T \phi(X) \phi(X)^T v + \lambda - \lambda v^T v .
\end{equation}
The GP prior $z \sim \mbox{GP}(0, k(\cdot, \cdot))$ is equivalent to a linear model in the Hilbert space $\Phi$, 
$$z(x) = \phi(x)^T v, \;\; v \sim N(0, I).$$
Maximizing the generalized posterior, we have
$$\max_Z \tilde{p}(Z \mid X) =\max_Z p(Z) \exp\{ - \ell(Z)\}  = \max_v p(v) \exp\{ - \ell(\phi(X)^T v)\}$$
When $J = 1$, the covariance loss is equal to 0. Under the assumption that $\sum_{i=1}^N \phi(x_i) = 0$, we have $\overline{Z} = 0$. The loss function $\ell(\phi(X)^Tv) = \mbox{Var}(\phi(X)^T v)$, therefore,
\begin{align*}
    \max_v p(v) \exp\{ - \ell(\phi(X)^Tv)\} &= \max_v \exp \left\{ c - \frac{1}2 v^T v + \frac{c_V}{N} v^T\phi(X)\phi(X)^T v \right \} \\
    &= \max_v \left\{ c - \frac{1}2 v^T v + \frac{c_V}{N} v^T\phi(X)\phi(X)^T v \right \},
\end{align*}
where $c$ is the log normalizing constant of the Gaussian distribution.
If  $\lambda = \frac{1}{2 c_V}$, the objective becomes
$$\max_{v} \left\{ 2c \lambda - \lambda v^Tv + \frac{1}N v^T \phi(X) \phi(X)^T v  \right\}.$$
By comparison with \Cref{eqn:kPCA}, the optimizer of the GPSSL posterior is equal to the first component of kernel PCA, provided $c_V = \frac{1}{2\lambda^*}$ where $\lambda^*$ is the maximizer of \Cref{eqn:kPCA}

\end{proof}

In the more general case of $J>1$, the covariance term decorrelates the representations (to an extent determined by the covariance loss weight $c_C$). While we no longer have exact equivalency to kernel PCA, this covariance loss performs a similar, albeit softer, role to the $v^Tv=1$ constraint in kernel PCA, which forces the representations to be orthogonal and shrinks the covariance matrix of the representations to 0.

\subsection{Using the resulting distribution over representations in downstream tasks}

One important motivation for SSL methods is to construct ``general purpose'' representations that can be used as inputs for downstream tasks---in particular, when we have a large amount of unlabeled data with which to learn our representations, but few labeled data for our downstream task. How can the distributions over representations, $\tilde{p}(z^*|x^*)$, obtained using GPSSL be used for downstream tasks?

The most straightforward option is to extract the mean of the distribution and use it analogously to representations obtained using deterministic SSL. We refer to this approach in our experiments as ``GPSSL-mean''.

While this approach may afford some robustness benefits over deterministic SSL methods, a downstream analysis based on such a point estimate of $f_z$ does not incorporate the uncertainty inherent in our representation learning.
One of the benefits of using GPSSL for representation learning is that it allows us to propagate uncertainty about our latent representation into our downstream task, as

$$\begin{aligned}p(Y|X) =& \int_\mathbb{R^{J\times N}}p(Y|Z)\tilde{p}(Z|X) dZ\\\approx& \sum_i p(Y|Z^{(i)}),\end{aligned}$$

where $Z^{(i)}\stackrel{\text{\small{iid}}}{\sim} \tilde{p}(Z|X)$. We refer to this approach in our experiments as ``GPSSL-full''. 
\section{Experiments}\label{sec:results}
To explore the performance of GPSSL, we first qualitatively look at the representations obtained on two synthetic datasets, before a quantitative analysis of the performance of GPSSL representations in downstream tasks. We further demonstrate the performance of GPSSL on relatively higher dimensional real datasets from UC Irvine Machine Learning Repository (UCI) repository, including Breast Cancer \citep{misc_breast}, Dermatology \citep{misc_dermatology_33}, Ecoli \citep{misc_ecoli_39}, and Mice Protein \citep{misc_mice_protein} dataset. We compare to VIGReg and kernel PCA, the two most similar methods. 

\subsection{Experimental settings}
Throughout, we set our representation dimension $J=5$. For GPSSL and kernel PCA we use a squared exponential kernel, 
$$k(x_i, x_j) = \sigma^2 \exp\left\{-\frac{1}{2}(x_i-x_j)^TL^{-2}(x_i-x_j)\right\},$$
where $L =\mbox{diag} (l_1, \dots, l_D)$ contains the lengthscale parameters. We fix $\sigma^2=1$, and set the lengthscale to be the maximum pairwise distance between observations and their $K$ nearest neighbors. We select the value of $K$, plus the weights $c_V$ and $c_C$ and the learning rate, based on the log-likelihood obtained on a validation set when using the representations to predict class labels based on the learned representations, using logistic regression. In the GPSSL variance loss $\varloss$ (Equation~\ref{eqn:varloss}) we fix $\gamma=1$ and $\eps=1e-7$.

We modify VICReg for use with tabular data. For both the encoder and the expander network, we use neural network with one hidden layer, with batch-normalization and ReLU activations. For the encoder, we use a hidden layer of width 10; for the expander, we use a hidden layer of size 5 (mirroring the size of our representations). We generate positive pairs by adding  Gaussian noise to the original data, whith standard deviation set to $r$ times the per-dimension dataset standard deviation. We select $r\in \{0.1, 0.2, 0.3\}$ and the learning rate based on validation set accuracy, as before.


\subsection{A qualitative look at GPSSL representations}\label{sec:qualitative}
\begin{figure}
\centering
\includegraphics[width=.5\columnwidth]{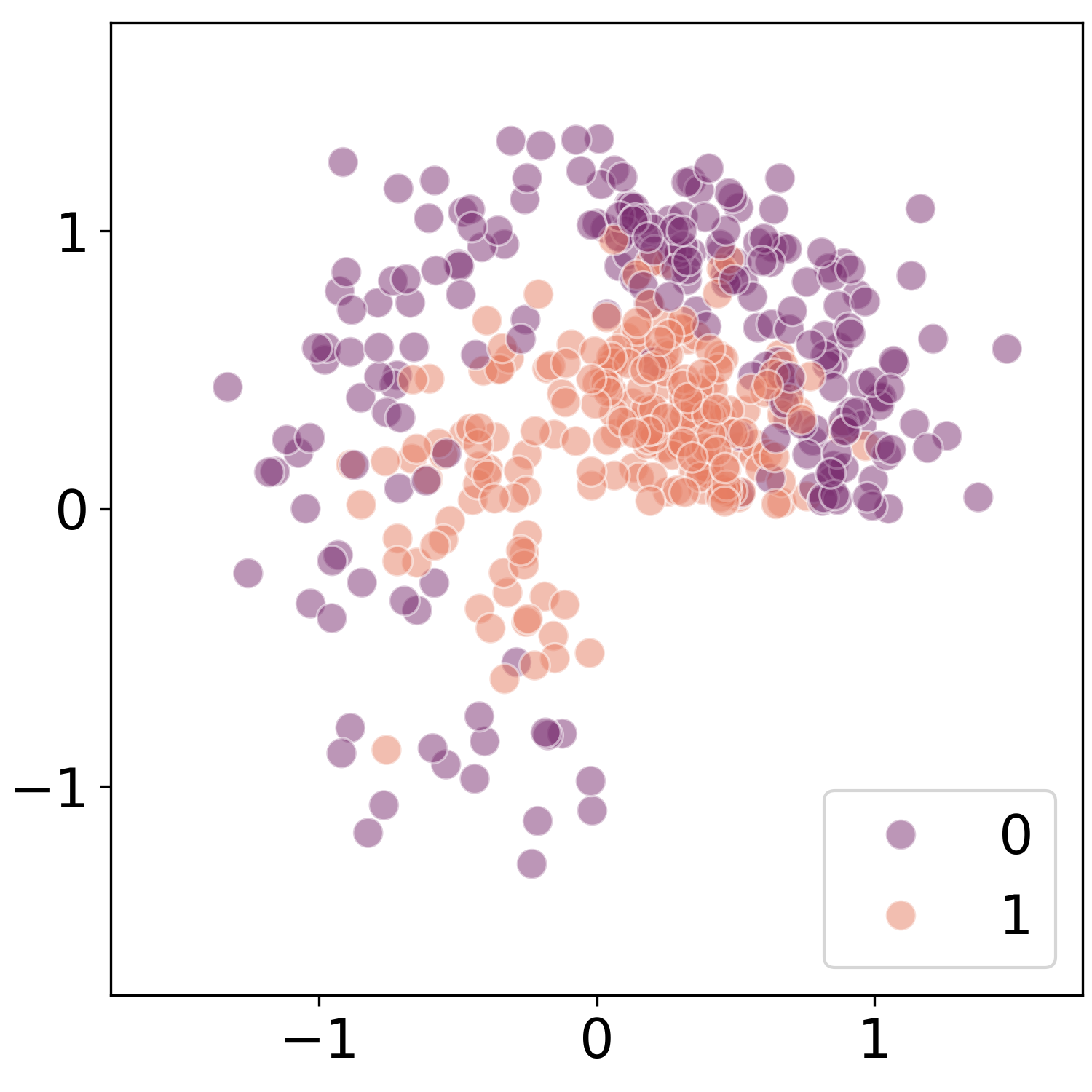}\caption{Quadrant-weighted concentric circles used to train embeddings. Labels are shown primarily for clarity; the training set labels are not used in training representations (but are used to select hyperparameters using a small validation set).}\label{fig:embedding_circle_train}
\end{figure}

We first consider learning representations for two-dimensional data generated from a known model. In order to explore how representation uncertainty varies with the prevalence of training data, we generate a dataset based on a quadrant-weighted pair of concentric circles.  We split the space into four quadrants, and generate 300 train/30 validation observations in the top right quadrant; 100 train/10 validation observations in the top-left quadrant; and 50 train/10 validation observations in the bottom-left quadrant. Within each quadrant, observations are sampled by selecting a radius from $\text{Uniform}\{0.5, 1\}$, uniformly selecting an angle within the quadrant, and adding Gaussian noise with standard deviation 0.2. The resulting dataset is shown in Figure~\ref{fig:embedding_circle_train}. 

For GPSSL, we fix $c_V=50$ and $c_C=10$. For the neighborhood size $k$ for GPSSL and kPCA, the learning rate for GPSSL and VICReg, the loss weights for VICReg, we use the validation set to select a value from a small grid. Concretely, a Bayesian logistic regression model is trained on 80\% of the validation observations to predict the originating circle (see \Cref{fig:embedding_circle_train}), to predict the originating circle, and the predictive log likelihood of the remaining 20\% is used for hyperparameter selection.

\begin{figure}[ht!]
\centering
    \begin{subfigure}{\textwidth}
    \centering
    \includegraphics[width=\columnwidth]{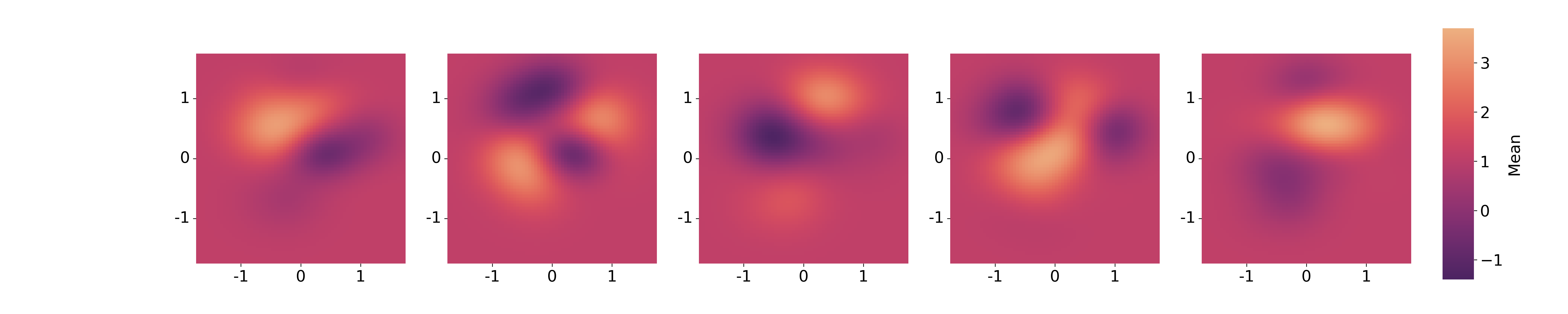}
        \caption{Mean of the representation function for each dimension.}\label{fig:mean_circ_gpssl}
    \end{subfigure}
    
    \begin{subfigure}{\textwidth}   
\includegraphics[width=\columnwidth]{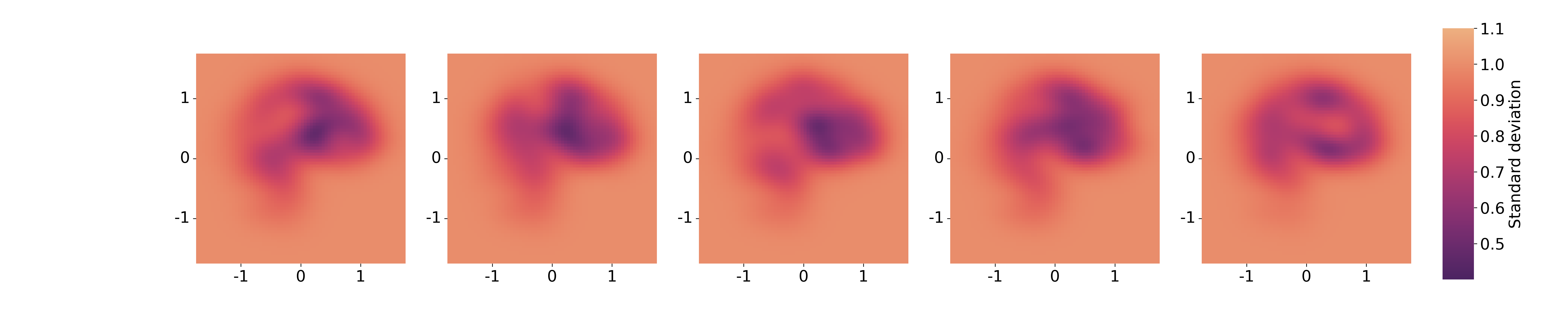}
        \caption{Standard deviation of the representation function for each dimension.}\label{fig:gpssl_circles_std}
    \end{subfigure}\\
    \centering
    \begin{subfigure}[t]{.32\textwidth}
    \centering
\includegraphics[width=\columnwidth]{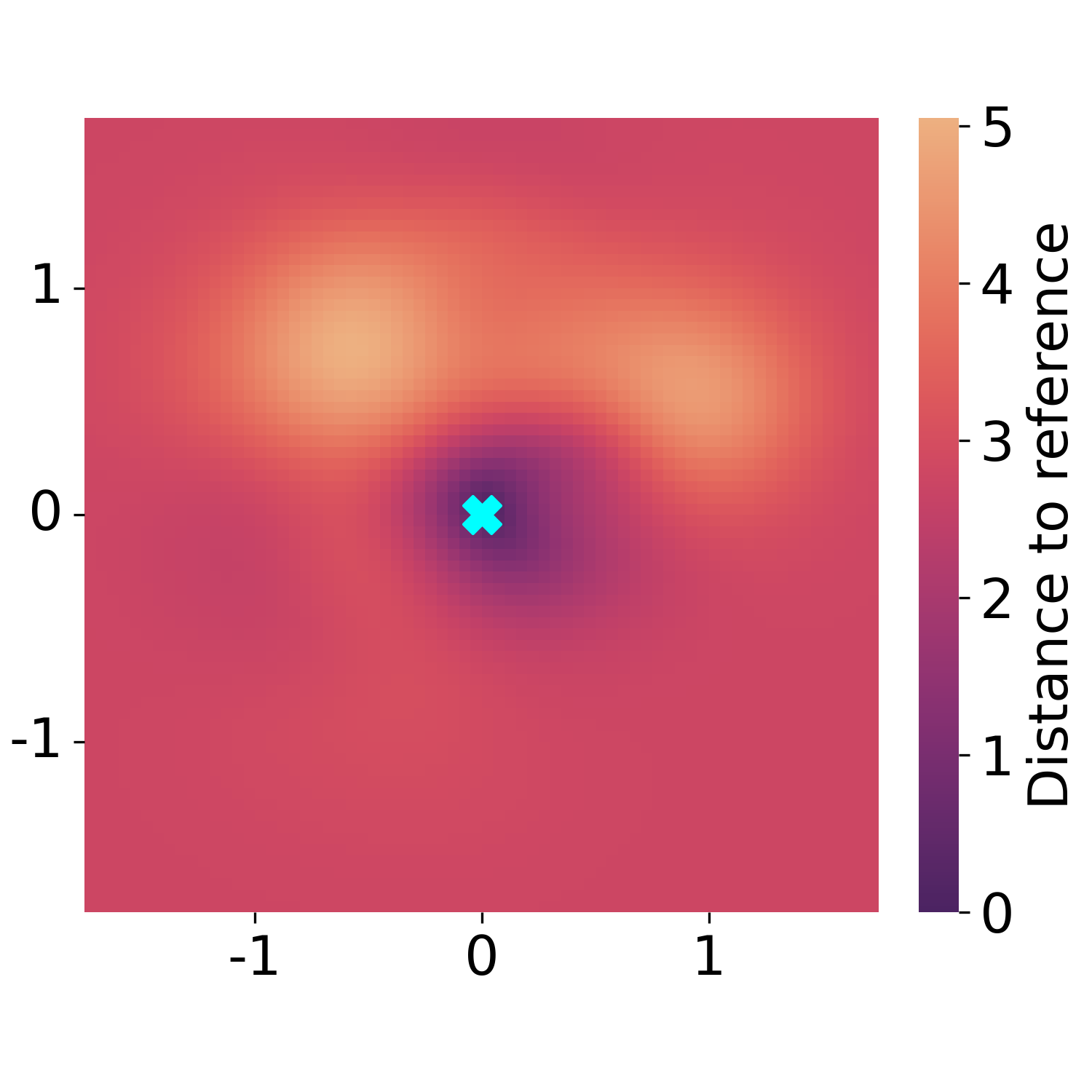}
        \caption{Distance  to reference.}\label{fig:ref_dist_circ}
    \end{subfigure}\qquad \qquad
    \begin{subfigure}[t]{.32\textwidth}
    \centering
\includegraphics[width=\columnwidth]{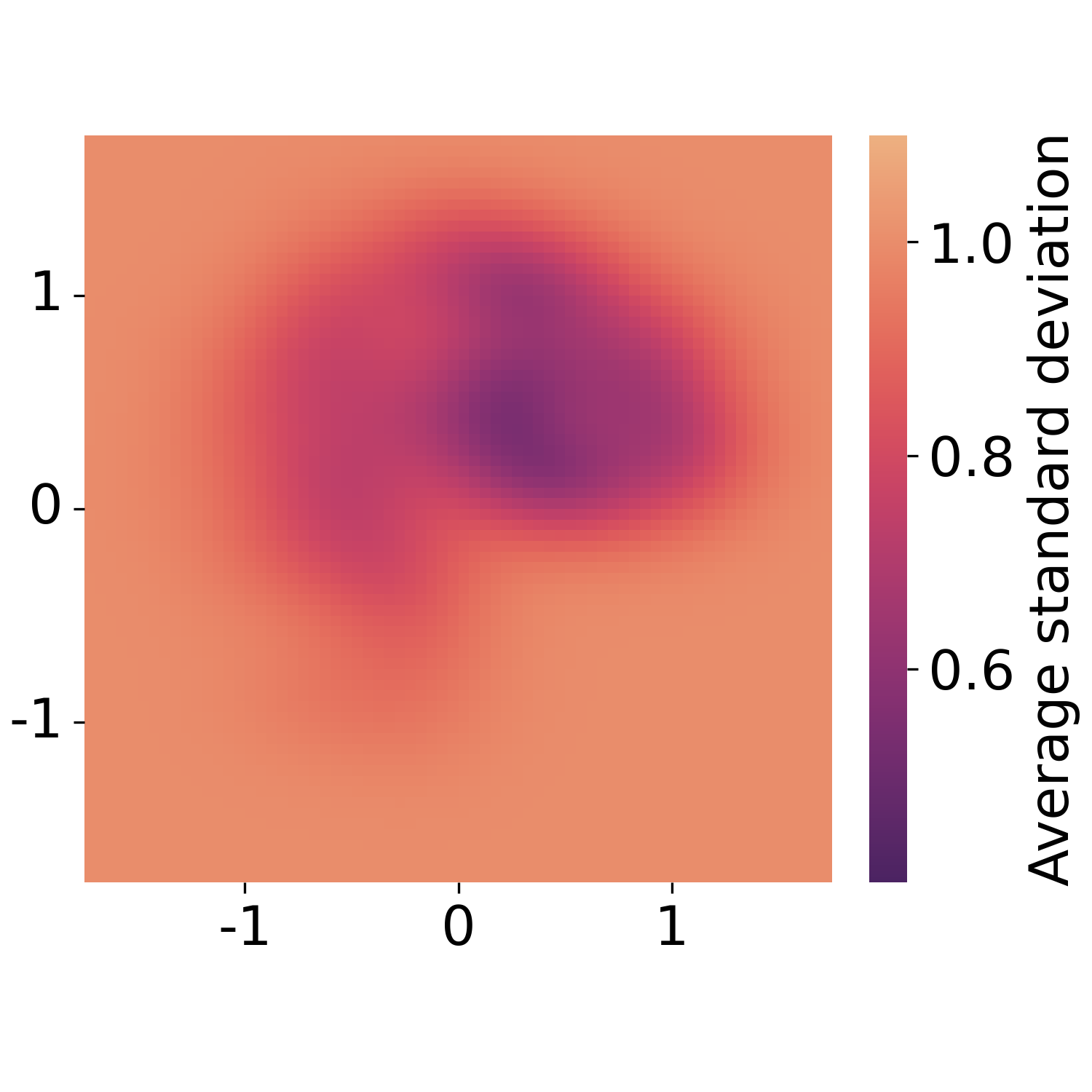}
        \caption{Average standard deviation.}\label{fig:gpssl_circles_std_average}
    \end{subfigure}
    
    \caption{Visualizations of GPSSL-based embedding function, trained on quadrant-weighted concentric circle data (\Cref{fig:embedding_circle_train}. Top two rows: Per-dimension mean and standard deviation. Bottom left: L2 distance between mean embedding at plot location, and mean embedding at (0, 0). Bottom right: average standard deviation (i.e., average of plots in \Cref{fig:gpssl_circles_std}).}\label{fig:gpssl_circle_embeddings}
\end{figure}

In~\Cref{fig:gpssl_circle_embeddings}, we show the embedding function learned using GPSSL. While the Gaussian process's kernels are aligned with Cartesian coordinates, it is able to combine dimensions to capture the circular structure (best shown in \Cref{fig:ref_dist_circ}, which shows the average L2 distances between the five-dimensional mean embeddings shown in \Cref{fig:mean_circ_gpssl}, and the mean embedding at (0, 0)).

We also see how the standard deviation of the embedding distribution varies across the space. In \Cref{fig:gpssl_circles_std}, we show that each dimension has lowest variance in the top right quadrant (where there were 300 training observations), with variance decreasing as we move counter-clockwise (as the number of training observations decreases to zero). This can be seen more clearly in \Cref{fig:gpssl_circles_std_average}, which shows the average of the five plots in \Cref{fig:gpssl_circles_std}. Also note that we have high variance as we move to the edges of the plot, away from the two concentric circles used in training; this is appropriate as we have no information about the distribution in these locations.

\begin{figure}[h!]
\centering
    \begin{subfigure}{\textwidth}
    \centering
    \includegraphics[width=\columnwidth]{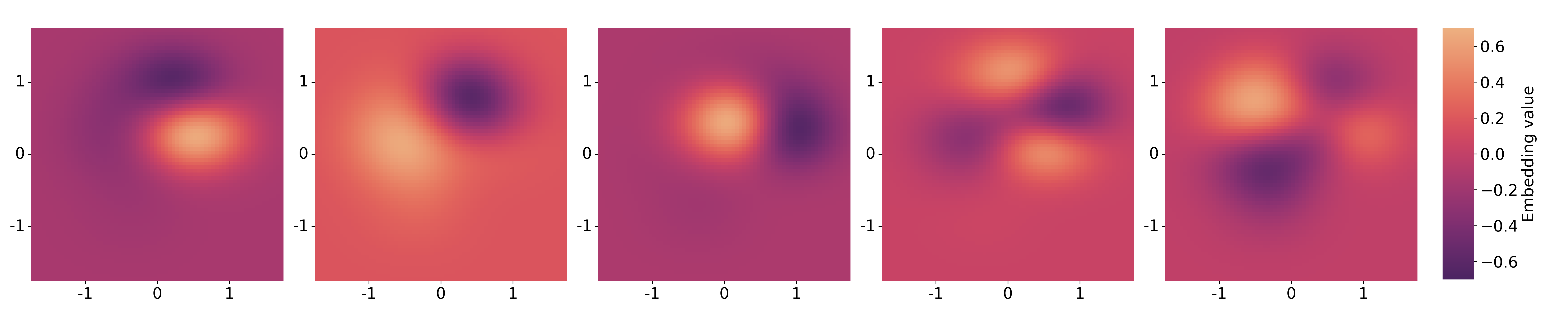}
        \caption{Five-dimensional representation obtained using kPCA}\label{fig:kpca_mean_circle_embeddings}
    \end{subfigure}\\
    \begin{subfigure}{\textwidth}
    \centering
    \includegraphics[width=\columnwidth]{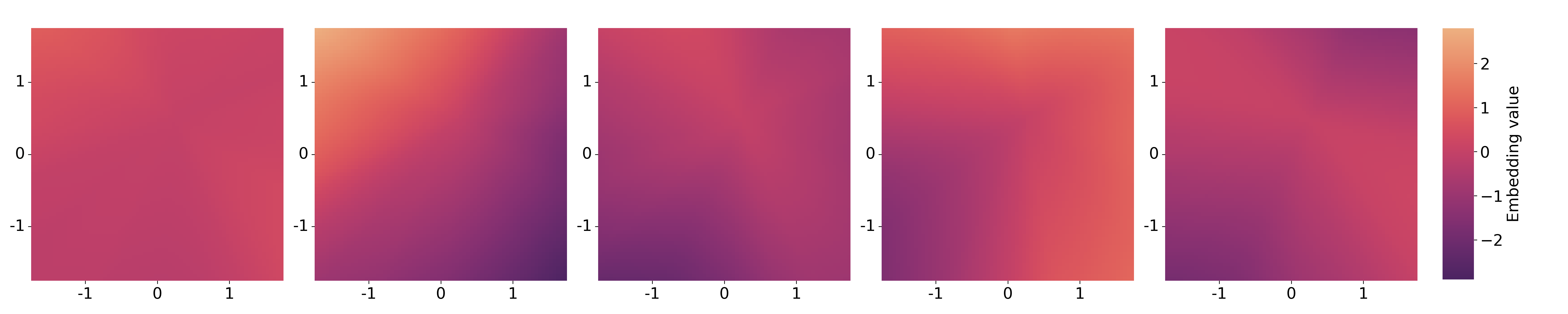}
        \caption{Five-dimensional representation obtained using VICReg}\label{fig:vicreg_mean_circle_embeddings}
    \end{subfigure}
    \begin{subfigure}{.35\textwidth}
    \centering
    \includegraphics[width=\columnwidth]{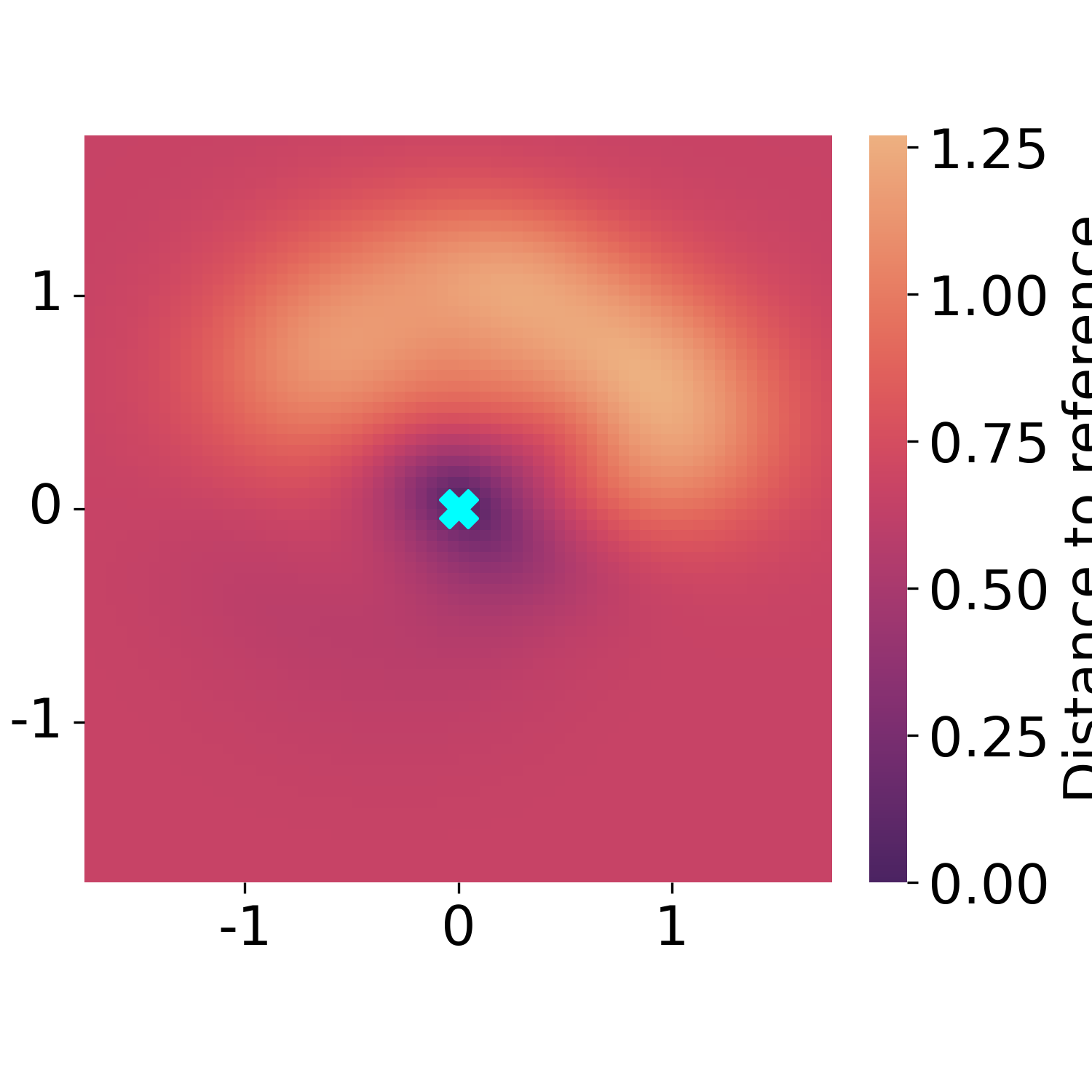}
        \caption{kPCA distance to reference.}\label{fig:kpca_ref_circle_embeddings}
    \end{subfigure}\qquad \qquad
    \begin{subfigure}{.35\textwidth}
    \centering
    \includegraphics[width=\columnwidth]{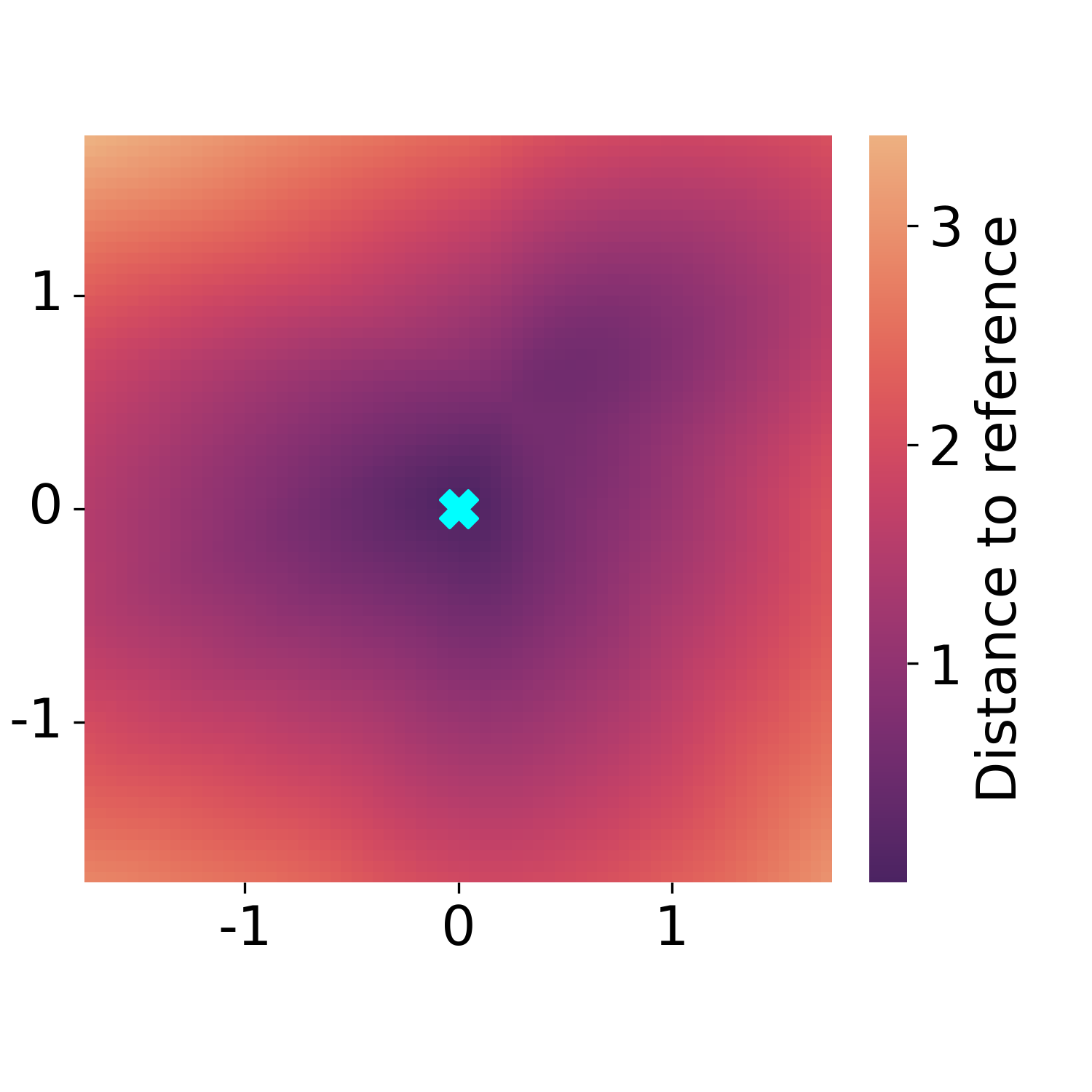}
        \caption{VICReg distance to reference.}\label{fig:vicreg_ref_circle_embeddings}
    \end{subfigure}\caption{Visualizations of kPCA-based and VICReg-based representation functions, trained on quadrant-weighted concentric circle data (\Cref{fig:embedding_circle_train}). Top row shows the five dimensions of the kPCA-based representation. Middle row shows the five dimensions of the VICReg-based representation. These are comparable with the GPSSL-based mean representation in \Cref{fig:mean_circ_gpssl}. Bottom row shows the L2 distance between the representation (left, kPCA; right, VICReg) at the plot location, and the representation at (0, 0). These are comparable with the GPSSL-based \Cref{fig:ref_dist_circ}.}\label{fig:kpca_vicreg_circle_embeddings}
\end{figure}


For comparison, \Cref{fig:kpca_mean_circle_embeddings} and \Cref{fig:kpca_ref_circle_embeddings} shows analogous embedding functions obtained using kPCA. The mean per-dimension embeddings (\Cref{fig:kpca_mean_circle_embeddings}), and the L2 distance to the embedding at (0, 0) (\Cref{fig:kpca_ref_circle_embeddings}), look similar to the plots for GPSSL in \Cref{fig:gpssl_circle_embeddings}. This is not surprising: Both approaches are based on the same kernels, and have the same mean representative power.  In \Cref{fig:vicreg_mean_circle_embeddings} and \Cref{fig:vicreg_ref_circle_embeddings}, we show the analogous embedding functions obtained using VICReg. Here, the representations obtained are more smoothly varying, but are still able to capture the radial structure of the data to some extent. 

While these representations seem comparable to the mean GPSSL representation (particularly in the case of kPCA), the kPCA and GPSSL representations \emph{do not} have any associated uncertainty, so we have no choice but to treat the embeddings in the bottom right quadrant (where there was no training data) as equally confident as the embeddings in the top right quadrant (which had the most training data).

\subsection{Evaluating (distributions over) representations for downstream tasks}

Next, we look at whether the representations learned using GPSSL are useful for downstream tasks. We begin with a qualitative exploration in \Cref{sec:dst_circles}, continuing the concentric circles example from \Cref{sec:qualitative}. Then, we explore quantitative performance on several tabular datasets in \Cref{sec:dst_UCI}.

\subsubsection{Qualitative evaluation of performance on downstream tasks}\label{sec:dst_circles}
\begin{figure}
\centering
    \begin{subfigure}{.45\textwidth}
    \centering
    \includegraphics[width=.8\columnwidth]{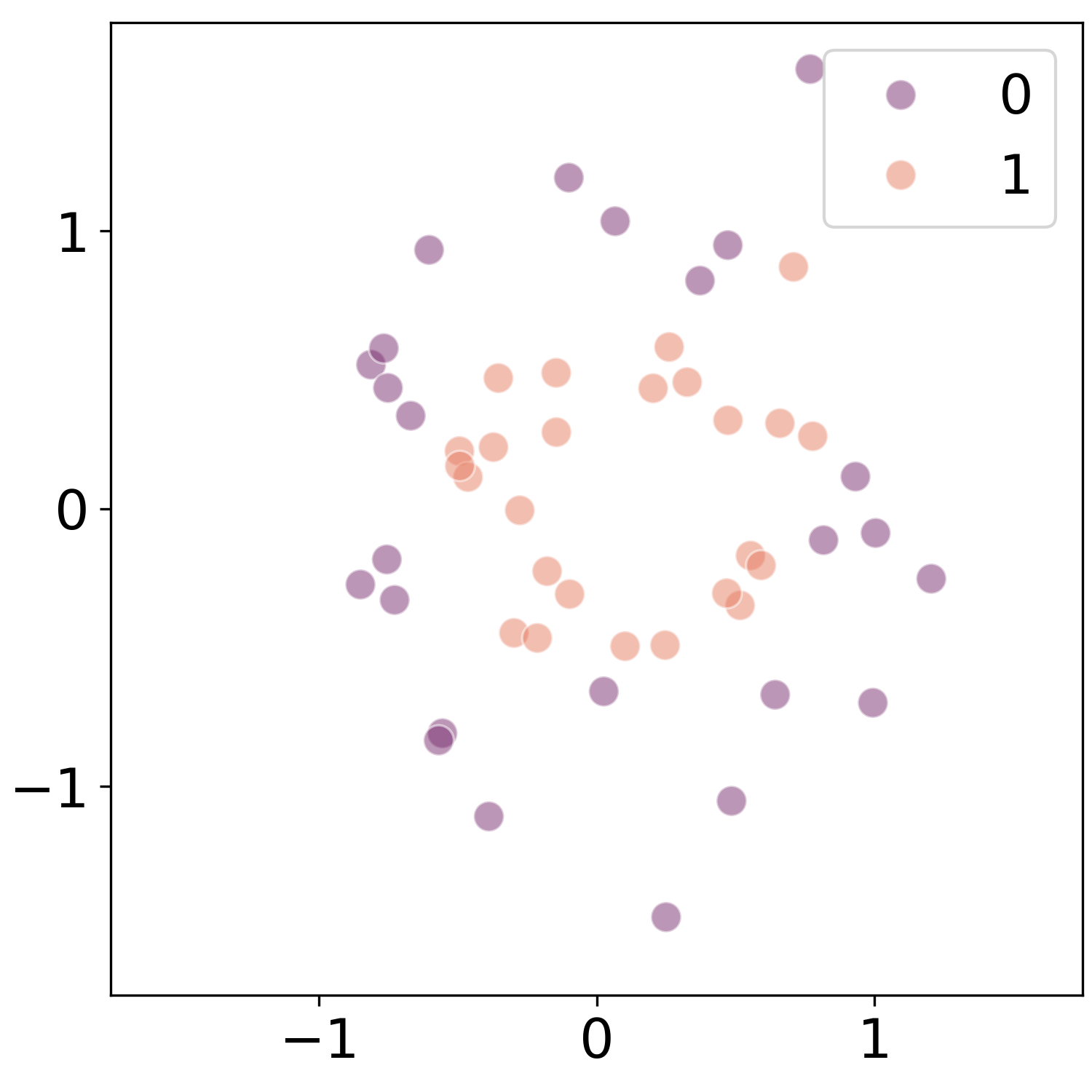}\label{Training set}\caption{Training data}
    \end{subfigure}
    \begin{subfigure}{.45\textwidth}
    \centering
    \includegraphics[width=.8\columnwidth]{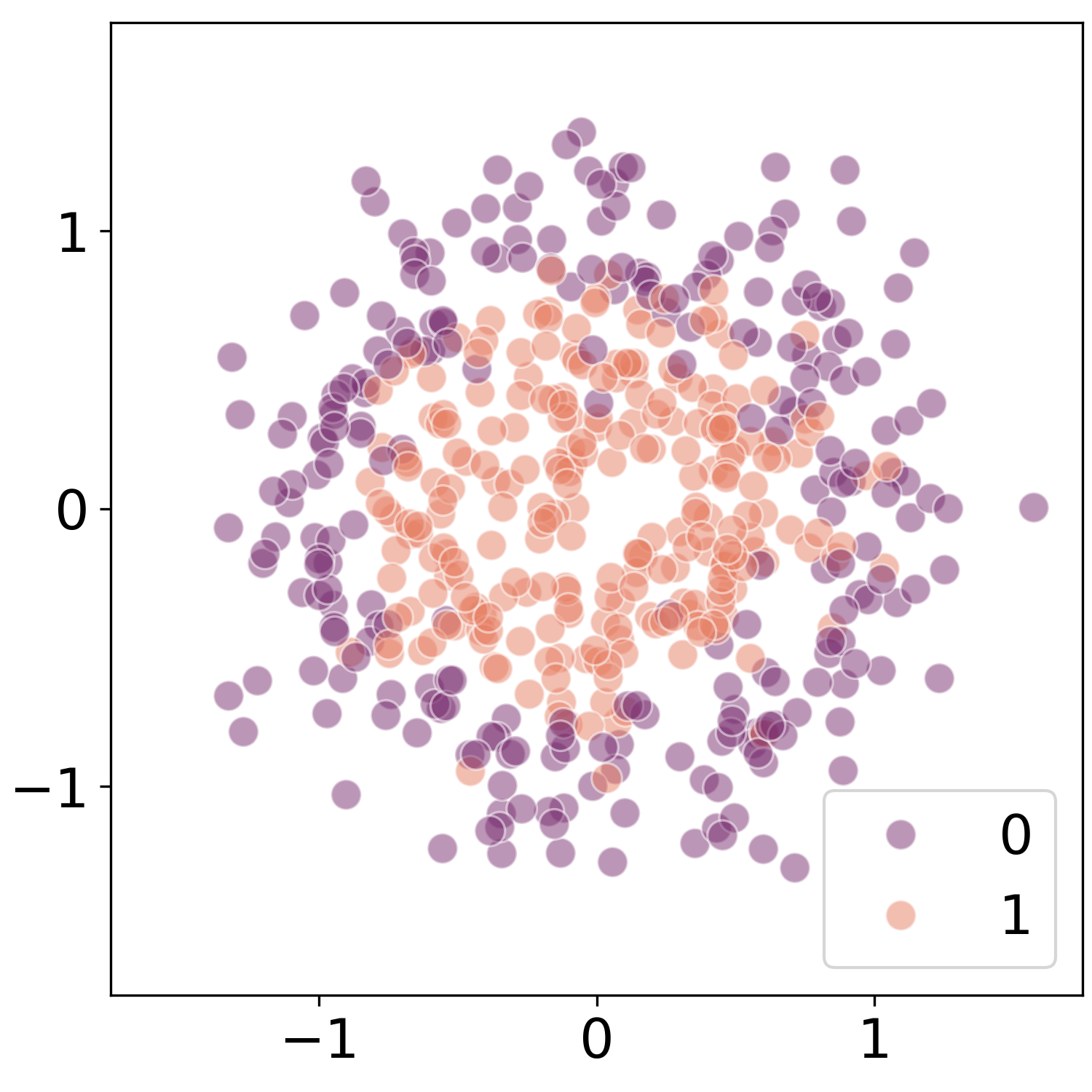}\label{Test set}\caption{Test data}
    \end{subfigure}\caption{Train and test set for concentric circles downstream classification task.}\label{fig:circles_dst_data}
\end{figure}

 Recall that in \Cref{sec:qualitative}, we learned a distribution over representations based on unevenly sampled data (\Cref{fig:embedding_circle_train}). We consider using this distribution over embeddings to build a classification model using a small number of observations sampled from a related distribution. Concretely, we sample 50 training observations and 500 test observations from a distribution over (non-quadrant-weighted) concentric circles, with mean and standard deviation as in \Cref{sec:qualitative}. These datasets are shown in \Cref{fig:circles_dst_data}.

We learn a Bayesian logistic regression model using the embedded training set. We consider two versions: In GPSSL-mean, we use the mean embedding as our covariates. In GPSSL-full, we treat the covariates as a random variable, and marginalize over their distribution under the GPSSL embedding distribution. We do this by generating 100 samples from the embedding distribution shown in \Cref{fig:gpssl_circle_embeddings} and learning a Bayesian logistic regression for each sample. 

\begin{figure}[h]
    \centering
    \includegraphics[width=\textwidth]{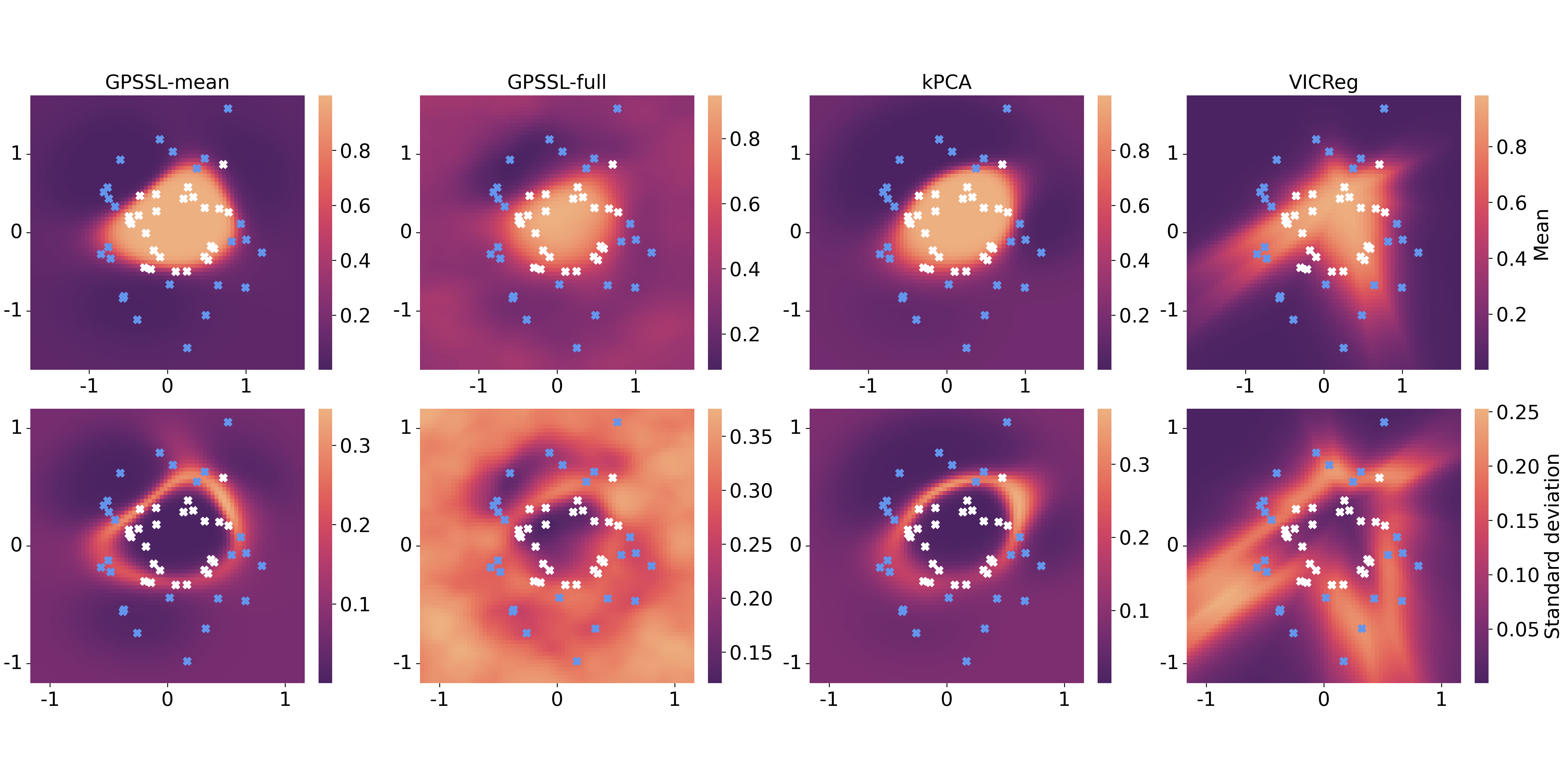}\label{fig:dst_smooth_probs}\caption{Mean (top) and standard deviation (bottom) of the predictive distributions obtained using Bayesian logistic regression on top of various representations of data (from left to right, GPSSL-mean, GPSSL-full, kPCA, VICReg).  Labeled training data is superimposed over each plot (white and blue crosses; color indicates class).}
\end{figure}

The resulting distributions are visualized in the first two columns of \Cref{fig:dst_smooth_probs}. We see that both learned classifiers have similar mean predictions (top row). And, because we have trained a Bayesian classification, both versions have associated uncertainties, even if trained on a point estimate of the embeddings (bottom row). However, making use of the full uncertainty in the embeddings leads to more meaningful uncertainty. If we look at the first column of \Cref{fig:dst_smooth_probs}, where we use just the mean of the GPSSL representation, we only have uncertainty around the boundary of the two classes. We remain confident in a prediction of class zero even far from the data. Conversely, the second column, where we use the full GPSSL distribution, we see much greater uncertainty as we move further from the data distribution. This reflects uncertainty in the underlying representation,as shown in \Cref{fig:gpssl_circles_std_average}.

The last two columns of \Cref{fig:dst_smooth_probs} show the distribution over class probabilities obtained using kPCA and VICReg-based representations. Since the kPCA representation is similar to the mean of the GPSSL representation (as discussed in \Cref{sec:qualitative}), the distribution over class probabilities for kPCA (third column) looks similar to the GPSSL-mean-based class probabilities. The mean prediction is reasonable, but we have unreasonably high confidence as we move away from the data distribution. The VICReg-based distribution (forth column) learns a much noisier classifier, a result of the smoother representations shown in \Cref{fig:vicreg_mean_circle_embeddings} and \Cref{fig:vicreg_ref_circle_embeddings}.

\begin{figure}[h]
    \centering
    \includegraphics[width=0.55\textwidth]{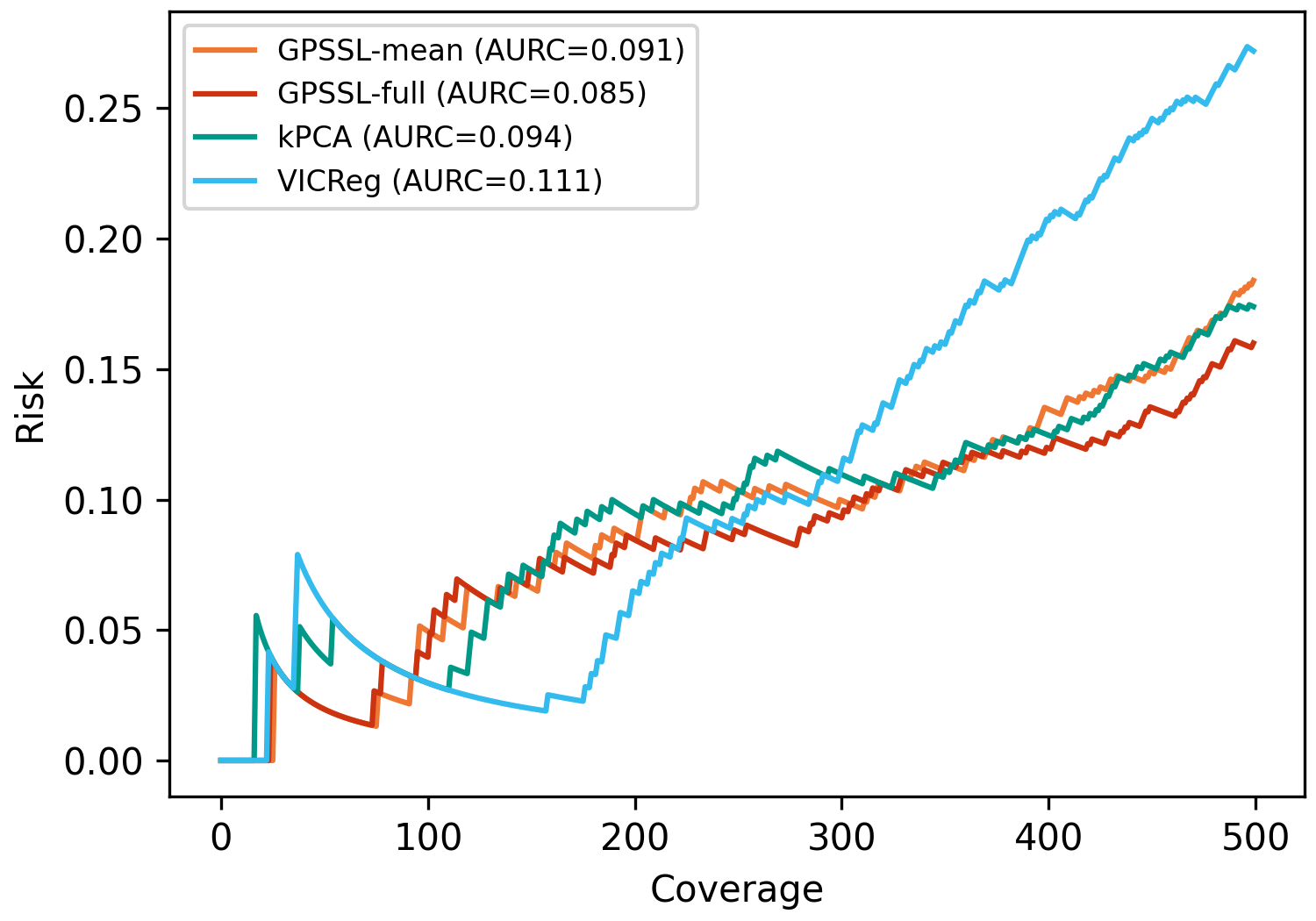}
    \caption{Risk coverage curve of GPSSL and the benchmark methods on circles data.}
    \label{fig:circles_GPSSL_rc}
\end{figure}

In \Cref{fig:circles_GPSSL_rc}, we show the risk-coverage curves of the mean function of the four classifiers, evaluated on the 500 test observations shown in \Cref{fig:circles_dst_data}. Risk-coverage curves evaluate performance on a ``selective classification'' task, where if the maximum class probability is above some threshold we assign that class, otherwise we decline to provide a class. The risk-coverage curve plots the risk (i.e., the proportion of data points above the classification threshold that are given erroneous classifications) vs the coverage (i.e., the proportion of data points that are above the classification threshold), as we vary the classification threshold. We see that the mean classifier obtained using GPSSL-full representations obtains the highest overall accuracy (i.e., the lowest risk at 100\% coverage), and has the lowest area under the risk-coverage curve (AURC), demonstrating the best overall utility in a selective classification task.

\subsubsection{Quantitative evaluation of performance on downstream tasks}\label{sec:dst_UCI}
Next, we evaluate the performance of GPSSL on four real datasets obtained from the UCI repository: Breast cancer \citep{misc_breast}; Dermatology \citep{misc_dermatology_33}; Ecoli \citep{misc_ecoli_39}; and Mice protein \citep{misc_mice_protein}. 

In each case, we split the dataset into a training set (40\% of observations; labels not included), a validation set (20\% of observations; labels included), and a test set (40\% of observations; labels to be predicted). We train our embedding models on the training set, using a grid of hyperparameter values as described in \Cref{app:additional_details}. We train a two-layer neural network classifier on 80\% of the validation set, and use performance on the remaining 20\% to select hyperparameters. We then train a classifier on the full validation set based on the corresponding representation function, and use it to predict the label on the test set

In \Cref{tab:classfy_UCI}, we look at quantitative performance on this test set. We report accuracy (based on a 0.5 threshold), area under the curve of a receiver operating characteristic (ROC AUC), and area under the risk-coverage curve (AURC, as described above). The best performance is highlighted in bold font. The classifiers trained on the full distribution from GPSSL (i.e., GPSSL-full) yields the best performance in most cases, and in other cases is comparable with the best-performing method. Of note, in all assessments, GPSSL-full outperforms GPSSL-mean (which ignores the variance of the representations), indicating the value of including uncertainty in representations.

\begin{table}[hbtp]
    \centering
    \begin{tabular}{ c|c|c|c|c|c } 
     \hline
     \hline
     \textbf{Breast Cancer} &  Original & VICReg & kernel PCA & GPSSL-mean & GPSSL-full \\ 
     \hline
     Accuracy & 0.965 & 0.891 & 0.969 & 0.956 & \textbf{0.974} \\ 
     ROC AUC  & 0.990 & 0.948 & 0.990 & 0.980 & \textbf{0.991} \\ 
     AURC & \textbf{0.005} & 0.029 & 0.007 & 0.009 & 0.006 \\ 
     \hline
     \textbf{Dermatology} &  Original & VICReg & kernel PCA & GPSSL-mean & GPSSL-full\\ 
     \hline
     Accuracy & 0.958 & 0.917 & \textbf{0.965} & 0.882 & \textbf{0.965}  \\ 
     ROC AUC & 0.997 & 0.986 & \textbf{0.996} & 0.987 & 0.994  \\ 
     AURC & \textbf{0.002} & 0.019 & 0.005 & 0.015 & 0.011  \\ 
     \hline
     \textbf{Ecoli} &  Original & VICReg & kernel PCA & GPSSL-mean & GPSSL-full\\ 
     \hline
     Accuracy & 0.859 & 0.726 & 0.822 & 0.822 & \textbf{0.881} \\ 
     ROC AUC & 0.894 & 0.858 & 0.896 & 0.893 & \textbf{0.906} \\ 
     AURC & 0.106 & 0.092 & 0.068 & 0.097 & \textbf{0.057} \\ 
     \hline
     \textbf{Mice Protein} &  Original & VICReg & kernel PCA & GPSSL-mean & GPSSL-full\\ 
     \hline
     Accuracy & 0.414 & 0.712 & 0.707 & 0.599 & \textbf{0.725} \\ 
     ROC AUC & 0.855 & 0.945 & 0.952 & 0.895 & \textbf{0.960} \\ 
     AURC & 0.630 & 0.166 & 0.156 & 0.294 & \textbf{0.078}  \\ 
     \hline
     \hline
    \end{tabular}
    \caption{Comparison of classification performance on UCI datasets.}
    \label{tab:classfy_UCI}
\end{table}

\newpage
\section{Real data application}

Next, we explore the use of GPSSL-based representations for analyzing spatial transcriptomics (ST) data. 
Spatial transciptomics is an RNA sequencing technique that measures gene expression 
across thousands of locations within an intact tissue slice, indexed by their two-dimensional spatial coordinates \citep{staahl2016visualization, zeira2022alignment}. 
Each location could possibly contain 10-100 cells with mixed cell types. We are interested in predicting the composition of cell types based on the spatially indexed RNA expression data.

\begin{figure}
    \centering
    \includegraphics[width=0.9\linewidth]{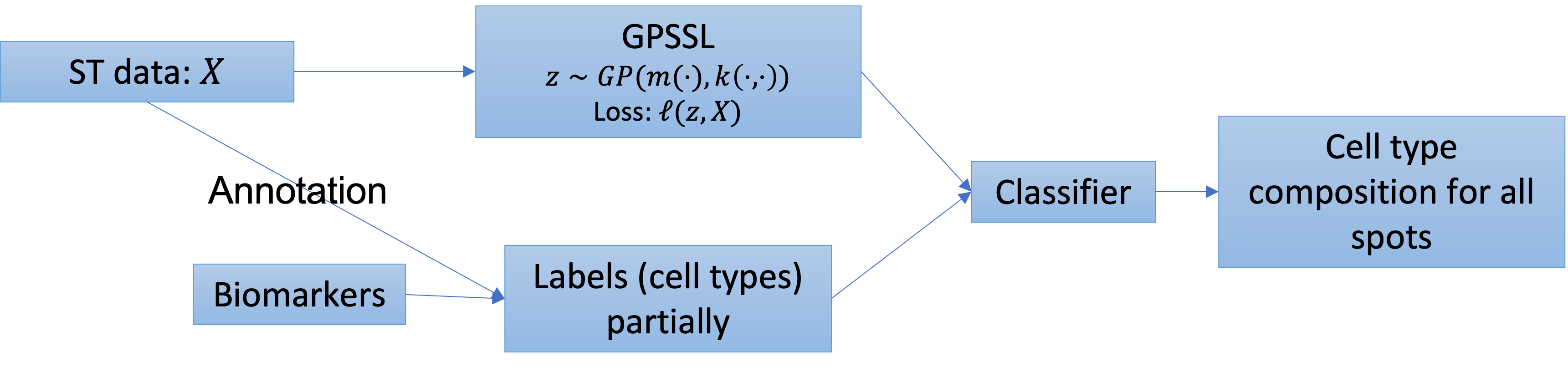}
    \caption{Pipeline of cell type identification on spatial transcriptomics data with GPSSL for representation learning.}
    \label{fig:STpipeline}
\end{figure}

Figure \ref{fig:STpipeline} shows our proposed pipeline, where we utilize GPSSL to learn representations for each of the spot considering all features observed from ST data.
We use Starfysh \citep{he2022starfysh} to identify the spots that are almost pure in cell type and annotate the cell types with known biomarkers. 
The analysis of cell type compositions for the rest of the spots could be considered as a classification problem, taking representations learned from GPSSL as features and cell types for the pure spots as labels. 
In the experiments, we use a Neural Network classifier, including two linear layers with ReLU as an activation in between.
A semi-simulation study is conducted to evaluate the performance of the proposed pipeline, and 
we apply this method on a breast cancer ST sample \citep{wu2021single} for cell type composition inference.

\subsection{Semi-synthetic study of cell type classification}\label{sec:semi-synthetic}

\begin{figure*}[htp]
    \centering
    \begin{subfigure}[b]{\textwidth}
        \centering
        \includegraphics[width = 0.8\textwidth]{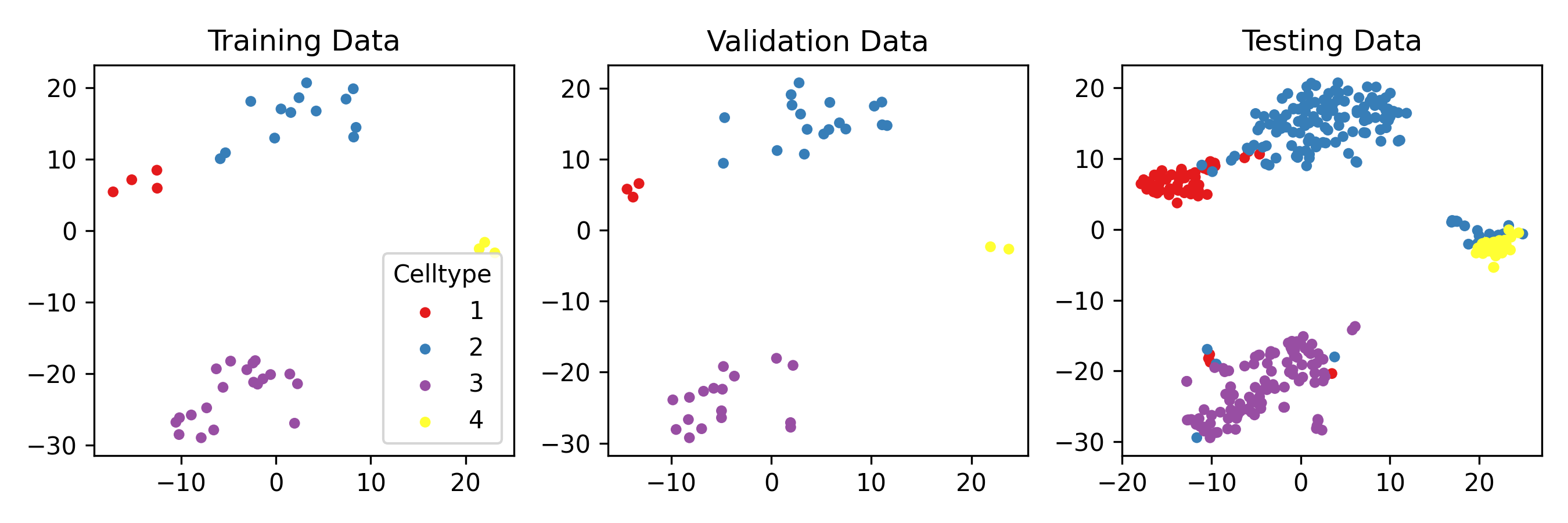}
    \end{subfigure}
    \begin{subfigure}[b]{\textwidth}
        \centering
        \includegraphics[width = 0.8\textwidth]{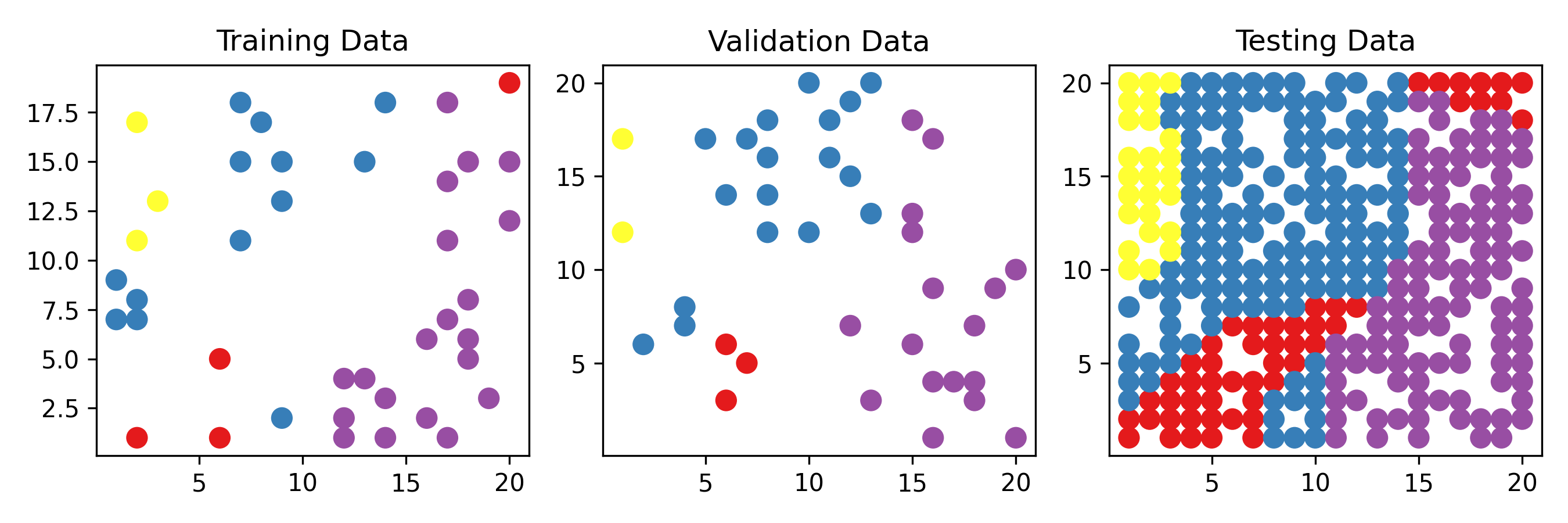}
    \end{subfigure}
    \caption{Simulated true data colored by the dominated cell type. The spots are plotted with respect to TSNE score of gene expression in the first row, and w.r.t. spatial coordinates on the tissue in the second row. }
    \label{fig:STsim_data}
\end{figure*}

Next, we design a semi-synthetic study to demonstrate the utility of GPSSL-based representations in ST data application. A simulated ST dataset is generated based on a Peripheral Blood Mononuclear Cells (PBMC) dataset, freely available from 10X Genomics \cite{pbmc}. We first generate high-fidelity
synthetic gene counts data with a simulator, scDesign2
\cite{sun2021scdesign2}, using the PBMC data as a reference. Four cell types including B cell, CD8 T cell, Memory CD4 T cell, and Naive CD4 T cell, in the PBMC dataset are used as ground truth in the semi-simulation study.
Then we simulate the proportions of each cell types across a grid of spots, based on the spatial coordinates. To simulate smoothly-varying cell-type allocations, we sample four Gaussian process traces, and use a softmax function to convert the values of these traces into a probability distribution at each location.  The simulated ST data is constructed by assigning single cells randomly based on the number of cells and the proportion of cell types in each spot. The cumulative gene counts of each spot is recorded as the ST data features.  \Cref{fig:STsim_data} shows the simulated data colored by the true celltypes, plotting with respect to gene expression TSNE scores and spatial coordinates in two rows, respectively.

\begin{figure}
    \centering
    \includegraphics[width=\linewidth]{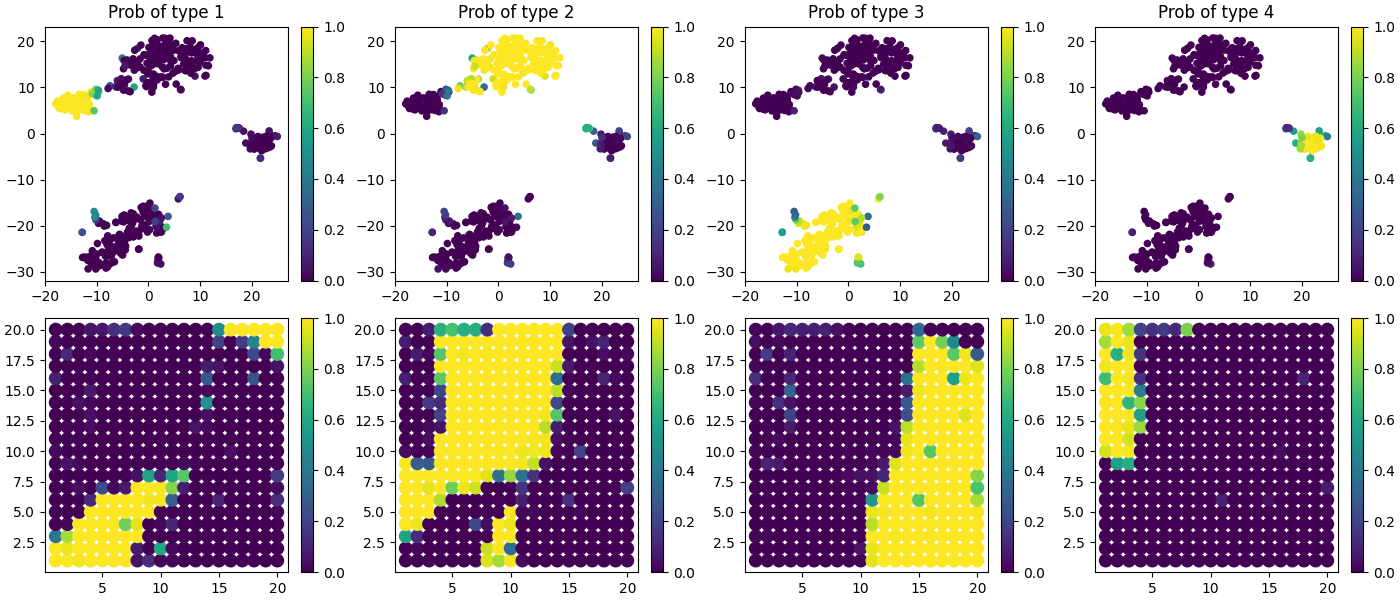}
    \caption{GPSSL result of estimating the probability of the four cell types. The two rows are plotted w.r.t. TSNE score of the gene expression and spatial coordinates, repectively.  }
    \label{fig:STsim_pest}
\end{figure}

We apply our proposed GPSSL pipeline to this semi-synthetic dataset, and look at the resulting cell type probabilities in \Cref{fig:STsim_pest}. Comparison with \Cref{fig:STsim_data} shows that we do a good job of predicting the appropriate cell type, and that the classifier is less certain on outlying observations. 

\begin{figure}
    \centering
    \includegraphics[width=0.6\linewidth]{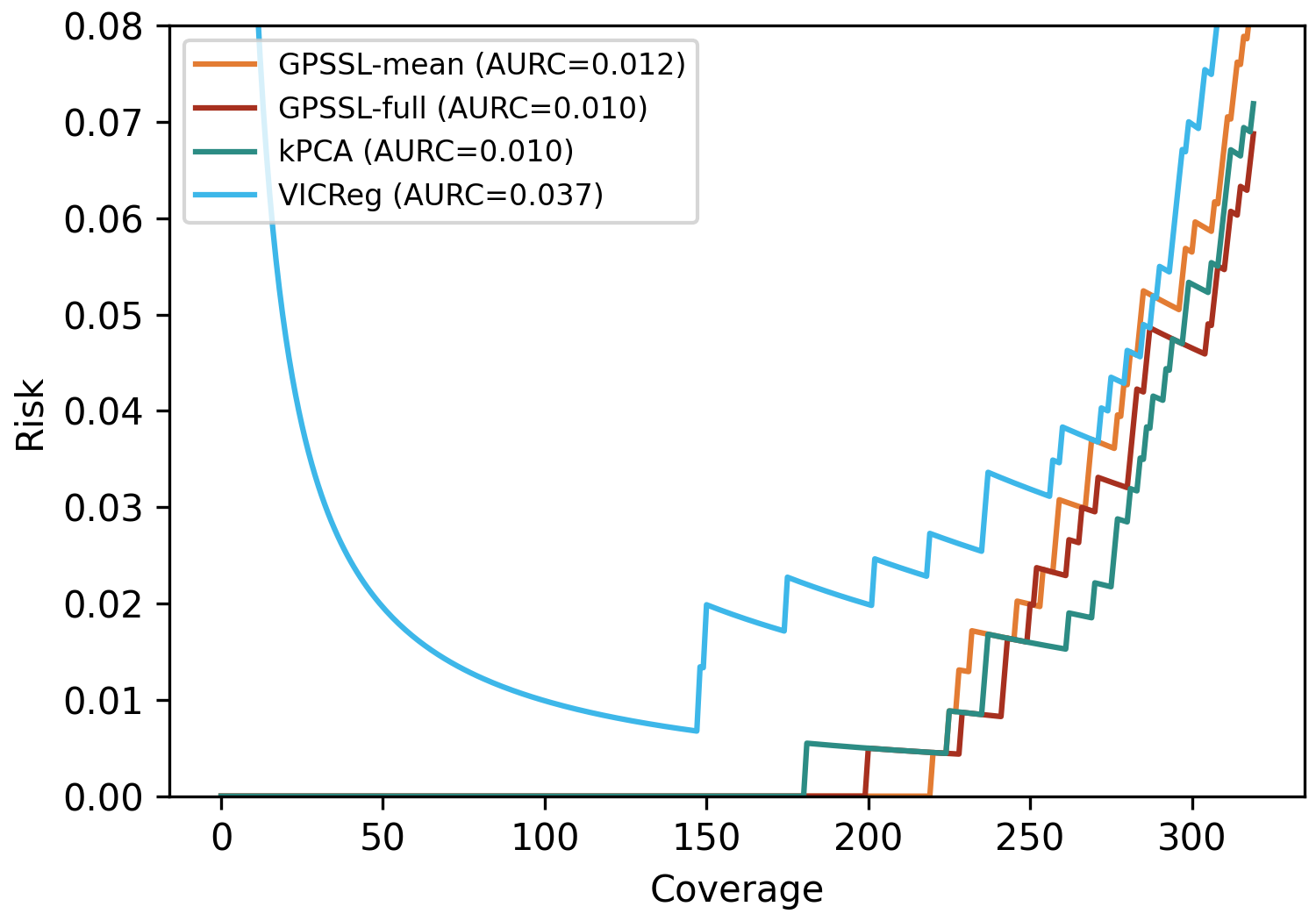}
    \caption{Risk coverage curves for the semi-synthetic ST dataset.}
    \label{fig:STsim_RC}
\end{figure}

Next, we compare performance with classifiers based on kPCA, VICReg, and the raw data. VICReg and kPCA are implemented as in previous subsections. \Cref{tab:classfy_STsim} shows the accuracy, ROC AUC, and AURC of the resulting classifier. Since we have access to the generating class probabilities, we also look at the mean squared error between the estimated (mean) class probabilities and the true probabilities (denoted pMSE). Here, we see that the full version of our GPSSL pipeline, and the mean-representation-based version of our pipeline, are consistently the best-performing methods, and that the full GPSSL pipeline performs best at modeling the true underlying probabilities. \Cref{fig:STsim_RC} shows the risk-coverage curve, where GPSSL yields a good proportion of coverage controlling the risk of mis-classification. 

\begin{table}[hbtp]
    \centering
    \begin{tabular}{ c|c|c|c|c|c } 
     \hline
     \hline
     \textbf{ST semi-synthetic} &  Original & VICReg & kernel PCA & GPSSL-mean & GPSSL-full \\ 
     \hline
     Accuracy & 0.922 & 0.906 & 0.928 & 0.919 & \textbf{0.931} \\
     ROC AUC  & 0.988 & 0.973 & \textbf{0.993} & 0.991 & 0.992 \\
     AURC & 0.029 & 0.037 & \textbf{0.010} & 0.012 & \textbf{0.010} \\
     pMSE & 0.021 & 0.027 & 0.019 & 0.027 & \textbf{0.014} \\
     \hline
     \hline
    \end{tabular}
    \caption{Evaluating test set predictions for GPSSL and competing methods, on a semi-synthetic ST dataset.}
    \label{tab:classfy_STsim} 
\end{table}

\subsection{An investigation of breast cancer cell-type composition}\label{sec:realdata}

\begin{figure}[htp]
    \centering
    \begin{subfigure}[t]{\textwidth}
        \centering
        \includegraphics[width=0.85\linewidth]{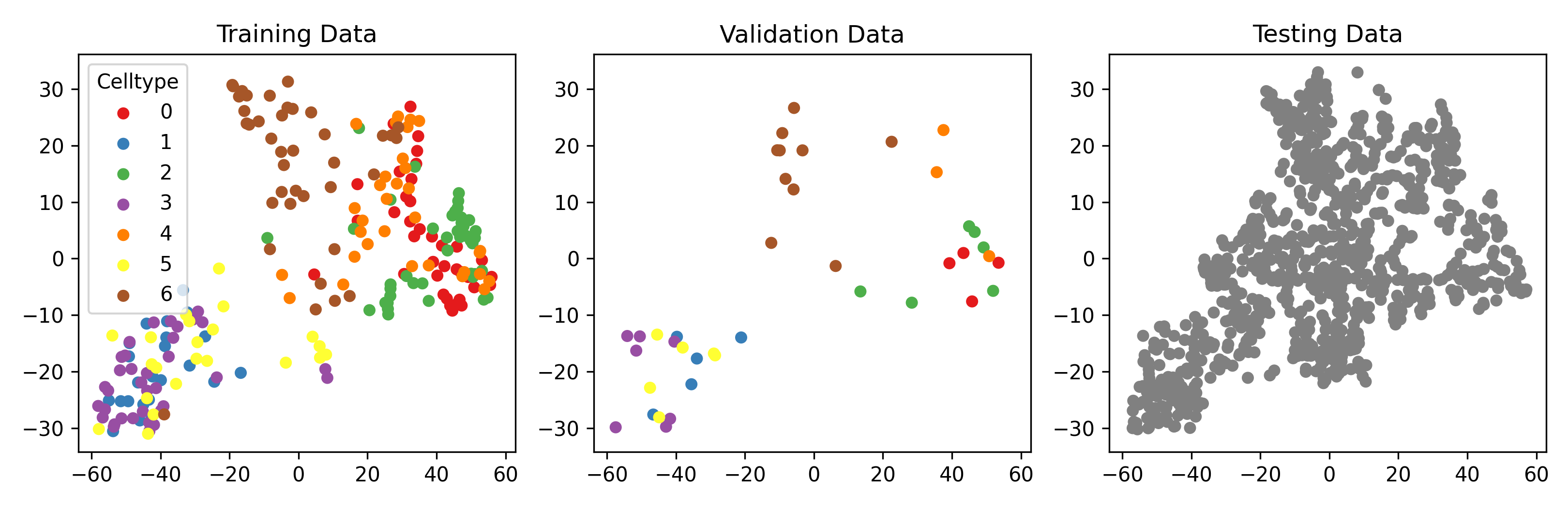}
    \end{subfigure}
    \begin{subfigure}[t]{\textwidth}
        \centering
        \includegraphics[width=0.85\linewidth]{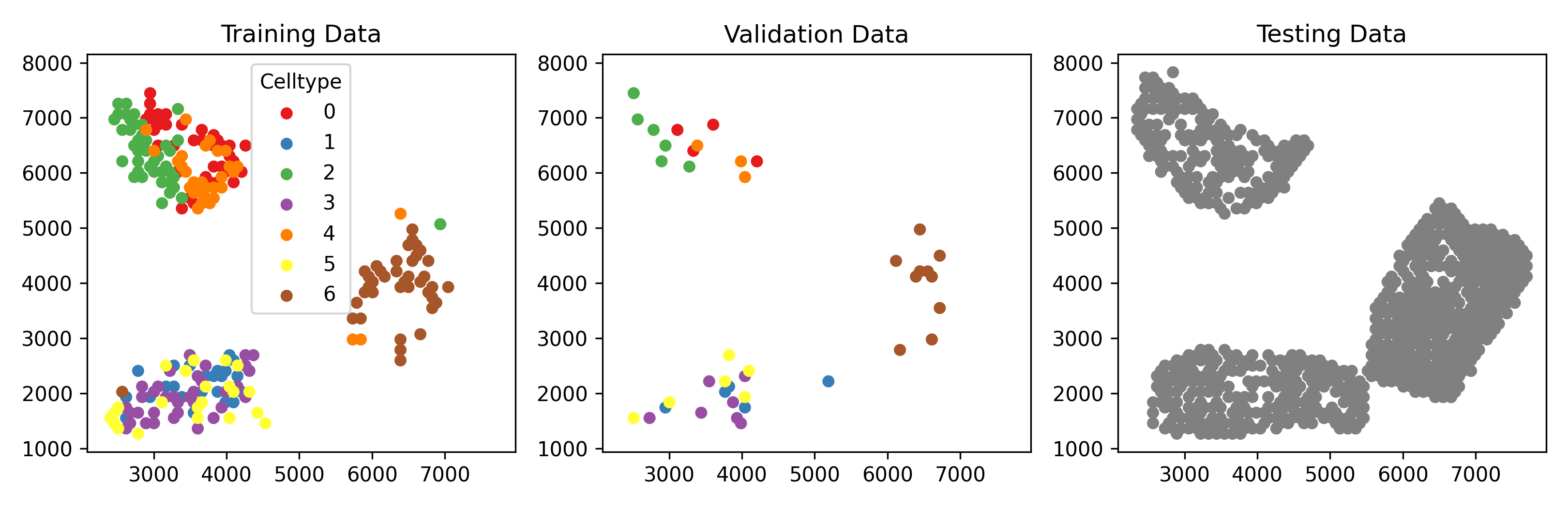}
    \end{subfigure}
    \caption{Breast Cancer data after preprocessing annotate with cell type labels 0-6. The color grey denotes the unlabeled data. The first row is plotted respect to TSNE score of the gene expressions, and the second row is plotted with respect to the spatial coordinates.}
    \label{fig:realdata_data}
\end{figure}

We investigate the cell type composition in a breast cancer sample (CID44971\_TNBC) from \citet{wu2021single} with RNA expression and spatial location observed at each spot, together with histology images recorded. We apply archetype analysis using the  Starfysh method \citep{he2024starfysh} to identify and annotate pure spots into different cell types. Figure \ref{fig:realdata_data} plots the breast cancer ST data after preprocessing, where labels 0-6 indicate identified cell types for the pure spots and  the label $-1$ denotes the unlabeled spots. We split the labeled data into training and validation set, and make inference on the cell types of the rest of the spots with the proposed pipeline. 

\begin{figure}
    \centering
    \begin{subfigure}[t]{\textwidth}
        \centering
        \includegraphics[width=\linewidth]{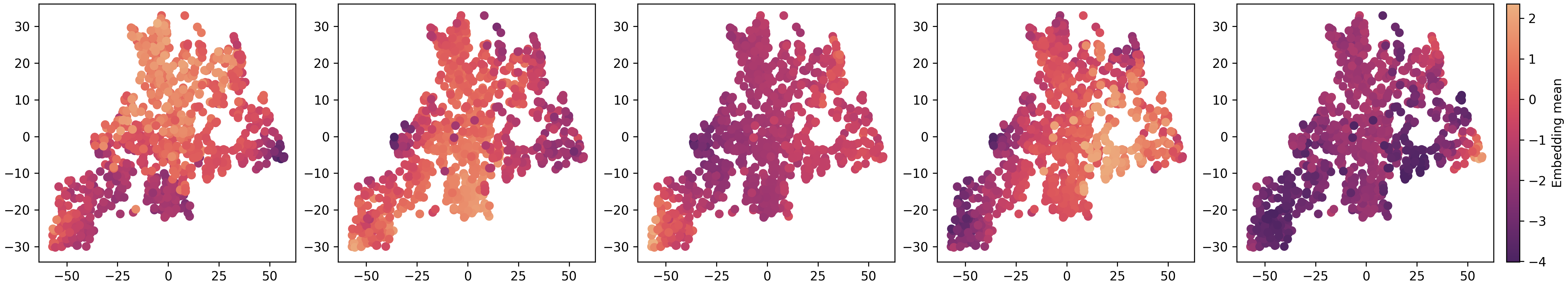}
        \caption{Mean of the representation for each dimension}
        \label{fig:realdata_repre}
    \end{subfigure}
    \begin{subfigure}[t]{\textwidth}
        \centering
        \includegraphics[width=\linewidth]{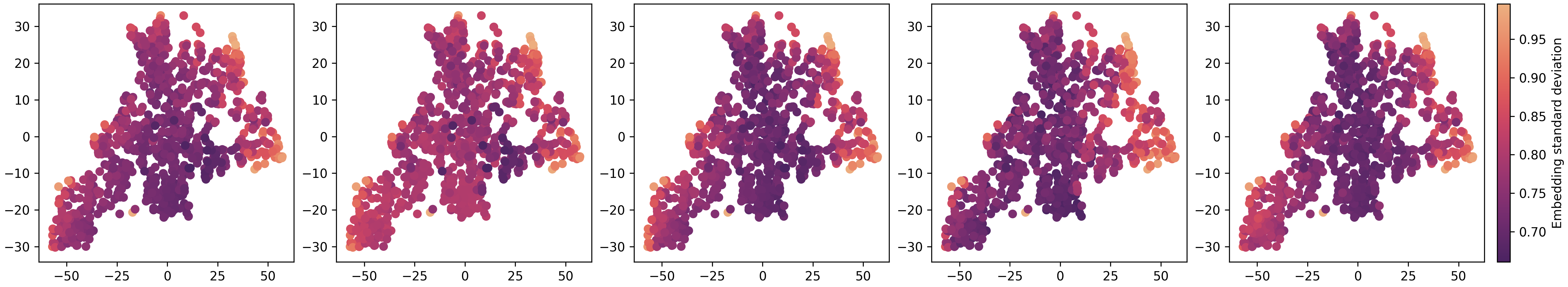}
         \caption{Standard deviation of the representation for each dimension}
         \label{fig:realdata_repre_sd}
    \end{subfigure}
    \caption{The learned representation and their standard deviations.}
    
\end{figure}

For computational reasons, we pre-process the gene expression data using PCA, and consider only the first 20 principle components. We combine these with the 2D spatial coordinates and the RGB values of the histology image to obtain input features for each spot. We then use GPSSL to learn a 5-dimensional distribution over representations, integrating the information from all three data features. \Cref{fig:realdata_repre} shows the five dimensions of the representations obtained, which reflect the structure of the data based on the features observed. The standard deviation of representations (\Cref{fig:realdata_repre_sd}) is a summary of the uncertainty in the representation learned, for example, on the edges of the data space where less data points are observed, we have relatively high uncertainty. This uncertainty will also be incorporated into the downstream analysis, in our case classification, and could serve as part of the assessment for the risk of mis-classification. 

In \Cref{fig:realdata_result}, we show the result of the entire embedding and classification pipeline. inference result on breast cancer data. The top row arranges the spots in terms of a TSNE embedding of the gene expressions; the bottom row arranges the spots based on spatial coordinates. In the first column, we see the highest-probability class predicted by our pipeline. We can see reasonable smoothness of the inferred cell types on both the gene expression and spatial location space. The middle column shows the standard deviation of this highest-probability class prediction. We see that the uncertainty of the dominant cell types increases when the spots are location in between different cell types or on the edge of the feature space. We can also report the probability of the spots being different cell types to convey a better understanding of the cell type composition, for example, the probability of being cell type 3 as is shown in the last column in \Cref{fig:realdata_result}.

\begin{figure}
    \centering
    \includegraphics[width=0.5\linewidth]{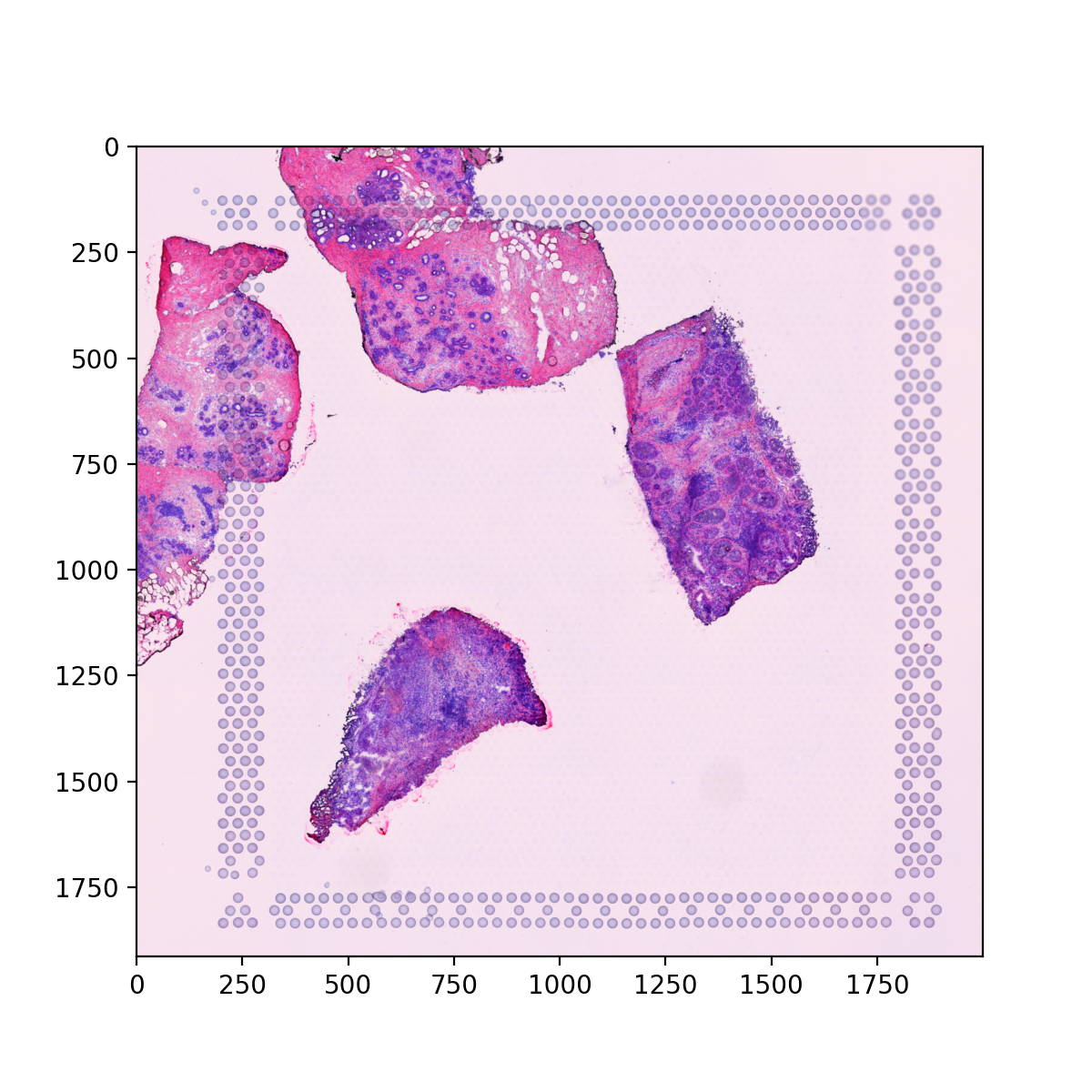}
    \caption{Histology image of the breast cancer sample.}
    \label{fig:realdata_histology}
\end{figure}

\begin{figure*}[htp]
    \centering
    \begin{subfigure}[t]{\textwidth}
        \centering
        \includegraphics[width=0.85\linewidth]{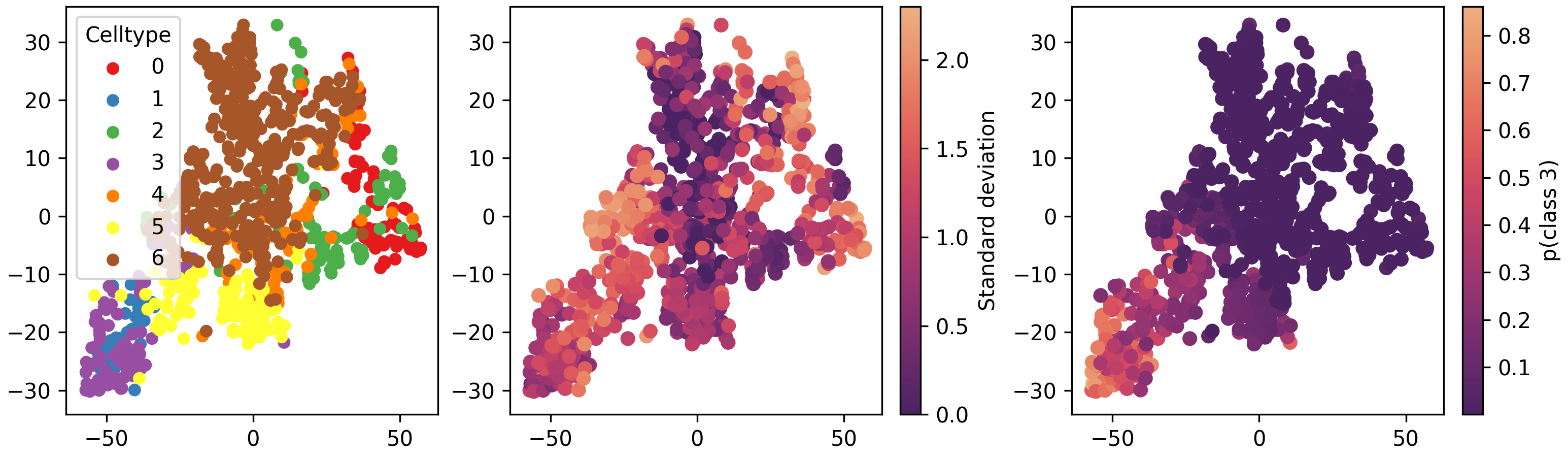}
    \end{subfigure}
    \begin{subfigure}[t]{\textwidth}
        \centering
        \includegraphics[width=0.85\linewidth]{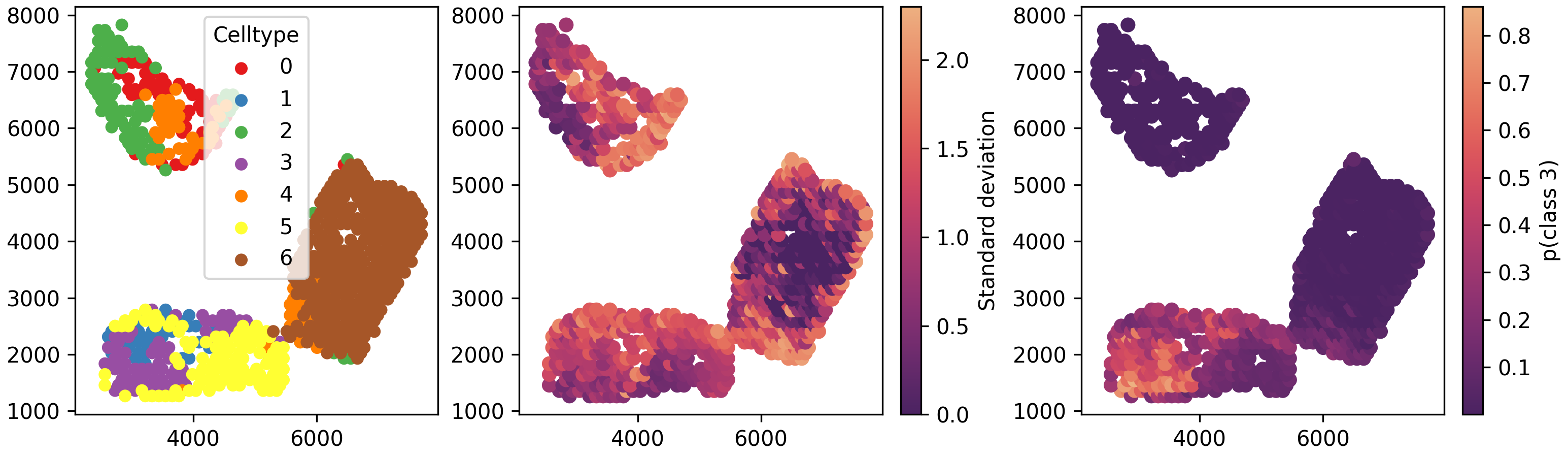}
    \end{subfigure}
    \caption{Analysis of breast cancer cell-type composition. Left hand column: highest probability class assignment. Middle column: standard deviation of the probability of the highest probability class. Right hand column: Probability of assignment to cell type 3. In the top row, cells are arranged based on a TSNE embedding of the gene expression. In the bottom row, cells are arranged based on spatial location (mirroring \Cref{fig:realdata_histology}).}
    \label{fig:realdata_result}
\end{figure*}

\section{Discussion}

We proposed Gaussian process self supervised learning that utilizes Gaussian processes on representation learning, allowing for full Bayesian inference and uncertainty quantification. We discussed the limitation of the current self supervised learning methods on certain data types and tasks, facing the difficulty of generating data pairs. It is demonstrated that the proposed GPSSL method tackles the problem by pulling representations of similar unit with the covariance function in GP. We discussed the connection between GPSSL and kernel methods and existing SSL approaches, and showed the benefits of using GPSSL in simulated and real data examples.

We specifically recommend this GPSSL approach on tabular, graph and other datasets with more structures. Other existing contrastive and noncontrastive learning methods might still be favorable considering the cases where the generation of positive or negative pairs of data is straightforward and highly informative. Under cases where the definition of similar objects is not easily available, GPSSL could characterize the similarity through the covariance function and the criterion could be tuned by the parameters in GPs. 

Though generalized variational inference is implemented for GPSSL and techniques like inducing points are used in GP regression, the computational feasibility of GPSSL is limited based on the sample size. It can become computationally impractical when dealing with a extremely large number of data points, due to the need to calculate and store a large covariance matrix.

The radial basis function (RBF) kernel is utilized in the proposed methods and implemented in the experiments. Extensions on different choices of the kernel in the GP model could be investigated. Although the RBF kernel is very widely used and should work well in most situations, a carefully crafted kernel will allow one to introduce for flexibility and effectively learn representations for certain datasets.

\section{Disclosure statement}\label{disclosure-statement}

The authors have no conflict of interest to declare.

\section{Data Availability Statement}\label{data-availability-statement}

Data used in experiments are publicly available at UC Irvine Machine Learning Repository at \url{https://archive.ics.uci.edu/}.
The public breast cancer dataset from \cite{wu2021single} can be accessed in the Gene Expression Omnibus under accession number GSE176078.  Processed data is available at Github repository created by \cite{he2022starfysh} at \url{https://github.com/azizilab/starfysh}.

\bibliography{bibfile}

\newpage
\appendix
\appendix

\section{Additional experimental settings}\label{app:additional_details}

For the experiments in \Cref{sec:results}, we explored different hyperparameter values for both GPSSL and our comparison methods, using a classification task on a validation set to select the best hyperparameters.

\textbf{Qualitative circle experiments (\cref{sec:qualitative})} For GPSSL, we varied the neighborhood size $K = \lfloor N/k \rfloor$ where $N$ denotes the sample size and $k \in \{20, 50, 100\}$ was used to select the kernel lengthscale, and the learning rate ($\ell \in \{0.01, 0.05, 0.001\}$). We fixed $c_V=50$ and $c_C=10$. For kPCA, we varied the neighborhood size $K = \lfloor N/k \rfloor$, $k \in \{20, 50, 100\}$ used to select the kernel lengthscale. For VICReg, we set $c_C=1$ and varied $c_V=C_I\in\{25, 50, 100\}$ (the original VICReg implementation used $c_V=c_I=25$ and $c_C=1$ \citep{bardes2021vicreg}), and varied the learning rate $\ell \in $ and the amount of noise added to the positive pairs (relative to the overall dataset set standard deviation), $\sigma\in\{0.1, 0.25, 0.5\}$. We set the learning rate to 0.001 for the first 20 iterations, then varied the learning rate $\ell \in \{0.00001, 0.00005, 0.0001, 0.0005\}$.

\textbf{UCI experiments (\cref{sec:dst_UCI}} )

For GPSSL, we varied the neighborhood size $K = \lfloor N/k \rfloor$ where $N$ denotes the sample size and $k \in \{5, 10, 20\}$ was used to select the kernel lengthscale, and the learning rate ($\ell \in \{0.01, 0.05, 0.001\}$). We fixed $c_V=50$ and $c_C=10$. For kPCA, we varied the neighborhood size $K = \lfloor N/k \rfloor$, $k \in \{5, 10, 20\}$ used to select the kernel lengthscale. For VICReg, we set $c_C=1$ and varied $c_V=C_I\in\{25, 50\}$ (the original VICReg implementation used $c_V=c_I=25$ and $c_C=1$ \citep{bardes2021vicreg}), and varied the learning rate $\ell \in $ and the amount of noise added to the positive pairs (relative to the overall dataset set standard deviation), $\sigma\in\{0.1, 0.25, 0.5\}$. We set the learning rate to 0.001 for the first 20 iterations, then varied the learning rate $\ell \in \{0.00001, 0.00005, 0.0001, 0.0005\}$.

\textbf{Semi-synthetic ST experiments (\Cref{sec:semi-synthetic})}

For GPSSL, we varied the neighborhood size $K = \lfloor N/k \rfloor$ where $N$ denotes the sample size and $k \in \{20, 50, 100\}$ was used to select the kernel lengthscale. We used learning rate ($\ell = 0.01$). We fixed $c_V=50$ and $c_C=10$. For kPCA, we varied the neighborhood size $K = \lfloor N/k \rfloor$, $k \in \{20, 50\}$ used to select the kernel lengthscale. For VICReg, we set $c_C=1$ and varied $c_V=C_I\in\{25, 50\}$ (the original VICReg implementation used $c_V=c_I=25$ and $c_C=1$ \citep{bardes2021vicreg}), and varied the learning rate $\ell \in $ and the amount of noise added to the positive pairs (relative to the overall dataset set standard deviation), $\sigma\in\{0.1, 0.25, 0.5\}$. We set the learning rate to 0.001 for the first 20 iterations, then varied the learning rate $\ell \in \{0.00001, 0.00005, 0.0001, 0.0005\}$.

\textbf{Breast cancer cell-type experiments (\Cref{sec:realdata})}

For GPSSL, we varied the neighborhood size $K = \lfloor N/k \rfloor$ where $N$ denotes the sample size and $k \in \{20, 50, 100\}$ was used to select the kernel lengthscale. We used learning rate ($\ell = 0.01$). We fixed $c_V=50$ and $c_C=10$.

Throughout, we trained GPSSL for 300 iterations and VICReg for 2000 iterations.

\clearpage \newpage

\section{Additional results on UCI datasets}


\begin{figure}[H]
    \centering
    \includegraphics*[width=0.8\textwidth]{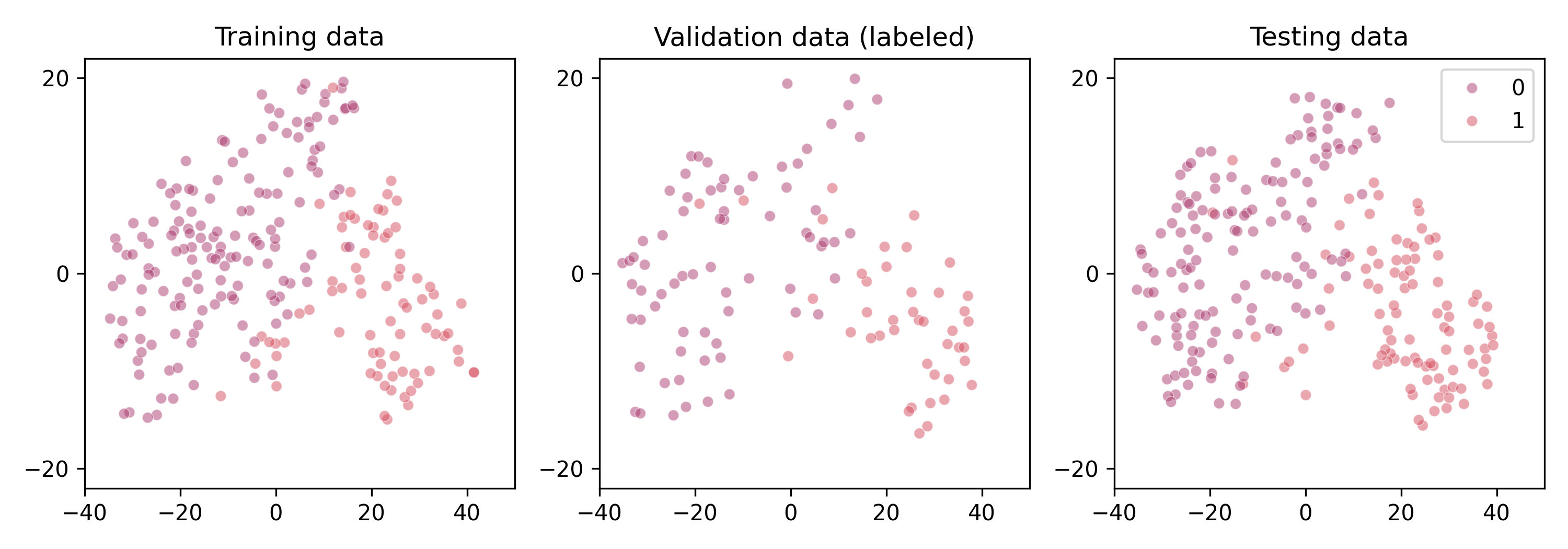}
    \caption{Breast cancer data in training, validation, and testing sets, plotted with respect to TSNE scores.}
    \label{fig:breast_data}
\end{figure}


\begin{figure}[H]
    \centering
    \includegraphics*[width=\textwidth]{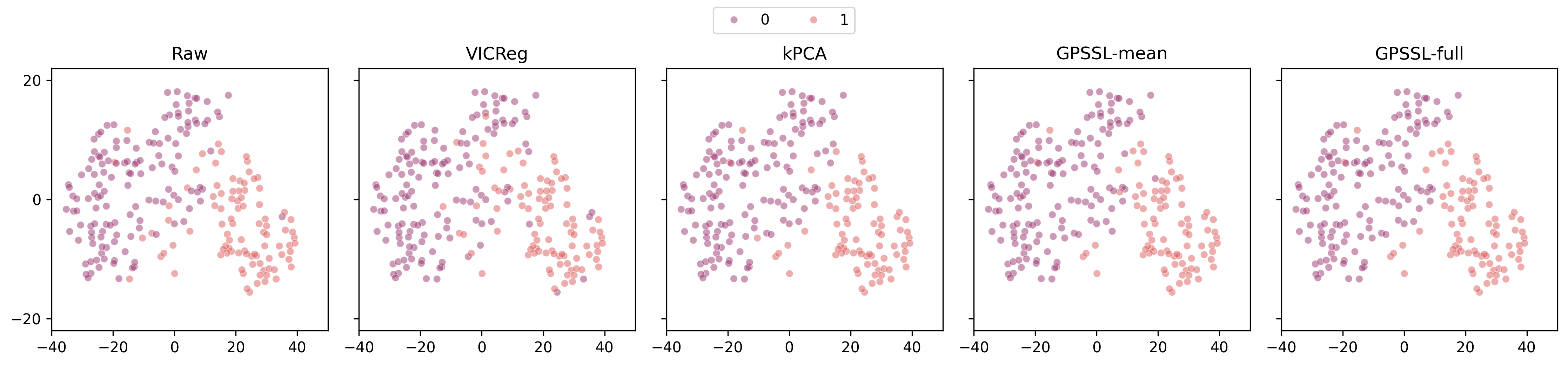}
    \caption{Classification results on breast cancer data, comparing original data, VICReg, kernel PCA, GPSSL-mean, and GPSSL-full.}
    \label{fig:breast_comparison}
\end{figure}

\begin{figure}[H]
    \centering
    \includegraphics*[width=0.45\textwidth]{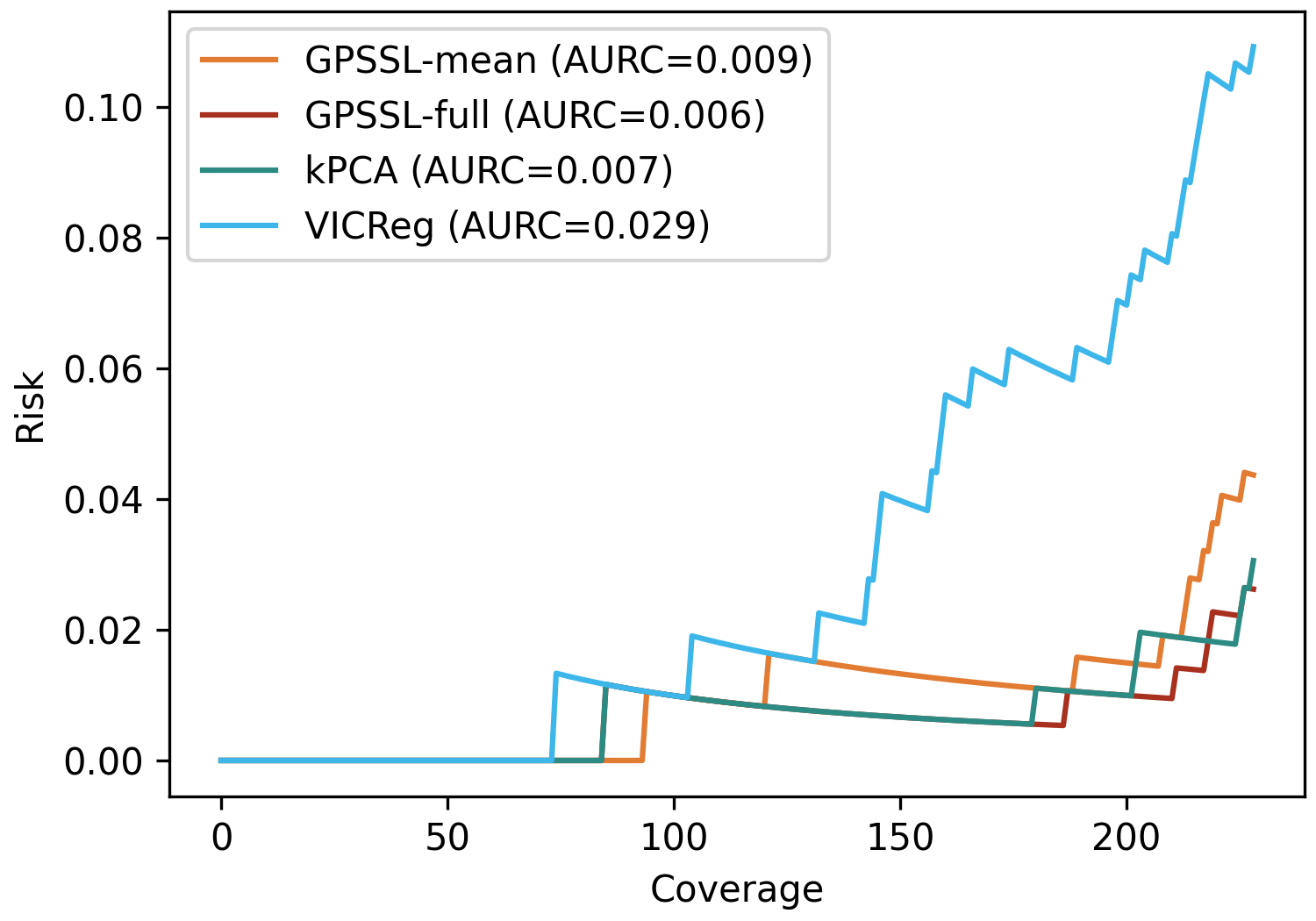}
    \caption{Risk coverage curve of GPSSL and the benchmark methods on breast cancer data.}
    \label{fig:breast_rc}
\end{figure}

\clearpage \newpage
\begin{figure}[H]
    \centering
    \includegraphics*[width=0.8\textwidth]{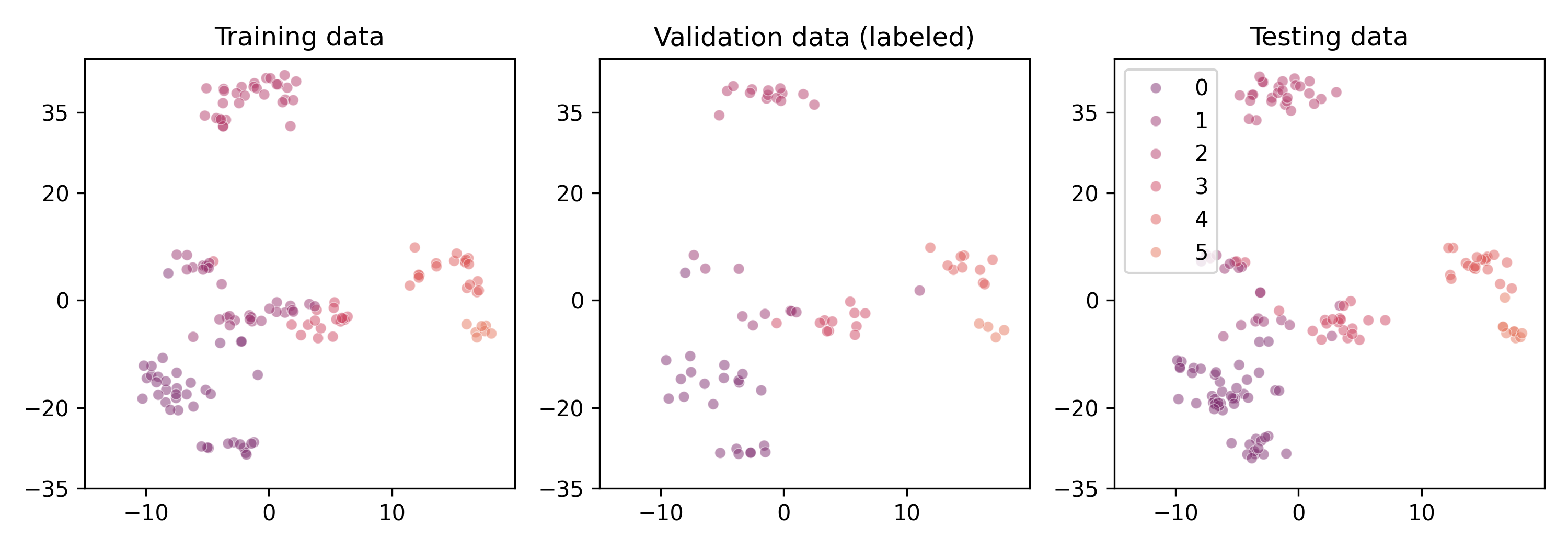}
    \caption{Dermatology data in training, validation, and testing sets, plotted with respect to TSNE scores..}
    \label{fig:derm_data}
\end{figure}

\begin{figure}[h]
    \centering
    \includegraphics*[width=\textwidth]{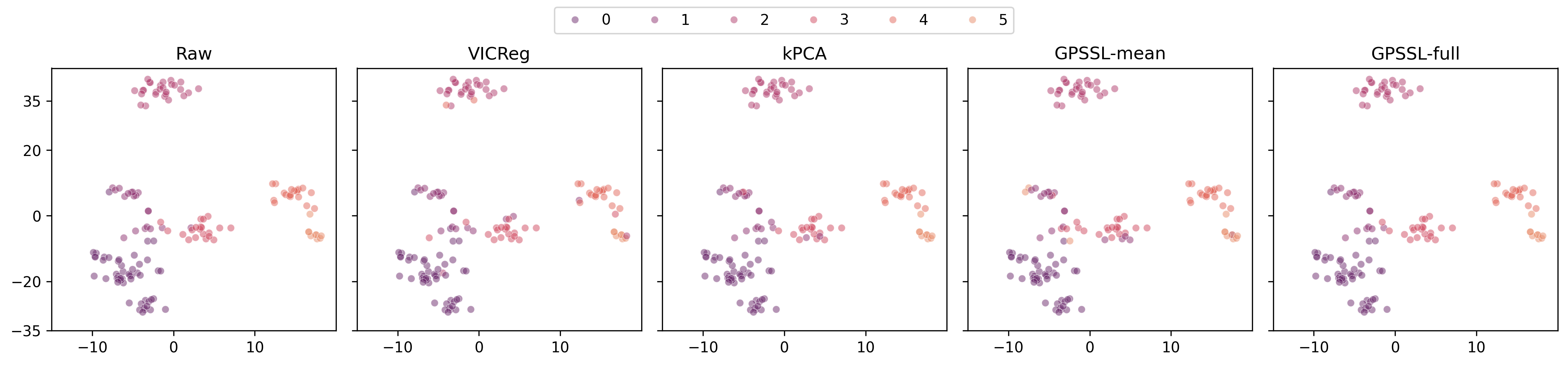}
    \caption{Classification results on dermatology data, comparing original data, kernel PCA, VICReg, GPSSL-mean, and GPSSL-full.}
    \label{fig:derm_comparison}
\end{figure}

\begin{figure}[h]
    \centering
    \includegraphics*[width=0.45\textwidth]{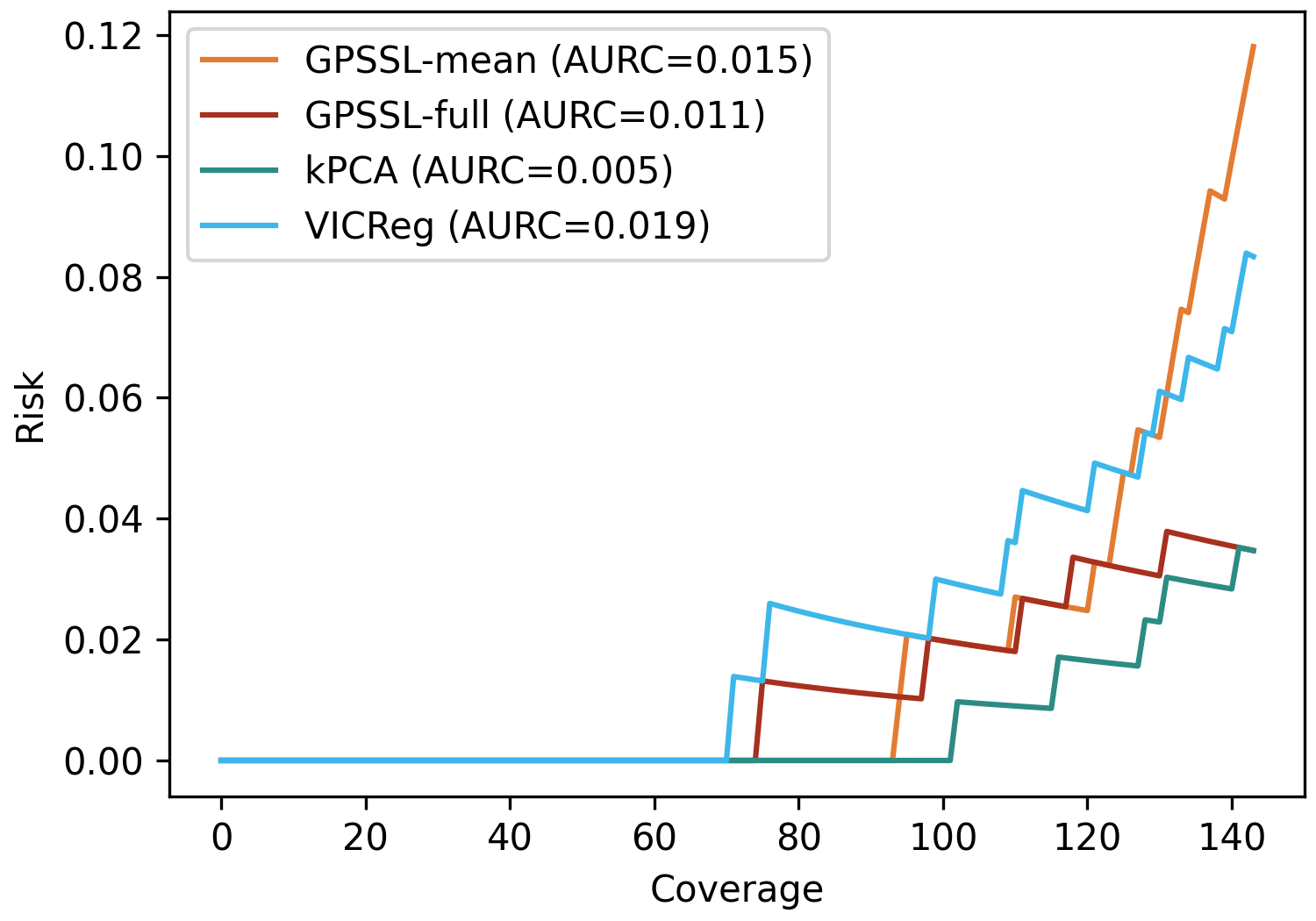}
    \caption{Risk coverage curve of GPSSL and the benchmark methods on dermatology data.}
    \label{fig:derm_rc}
\end{figure}

\clearpage \newpage

\begin{figure}[H]
    \centering
    \includegraphics*[width=0.8\textwidth]{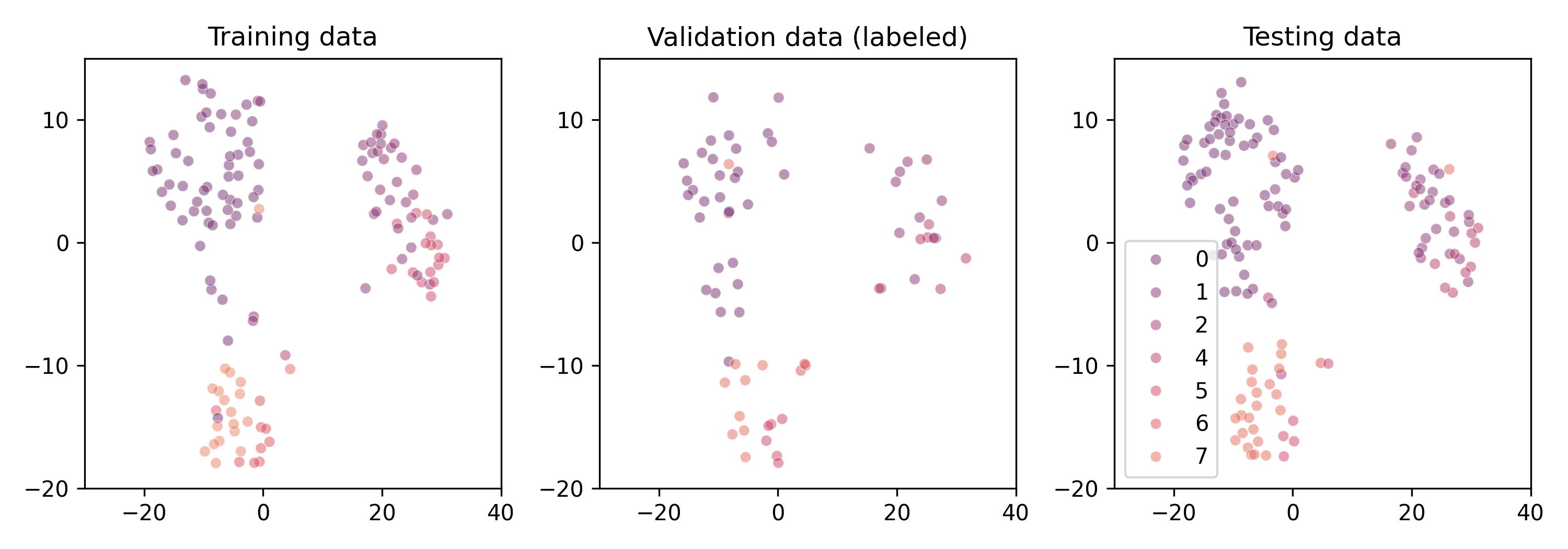}
    \caption{Ecoli data.}
    \label{fig:ecoli_data}
\end{figure}

\begin{figure}[h]
    \centering
    \includegraphics*[width=\textwidth]{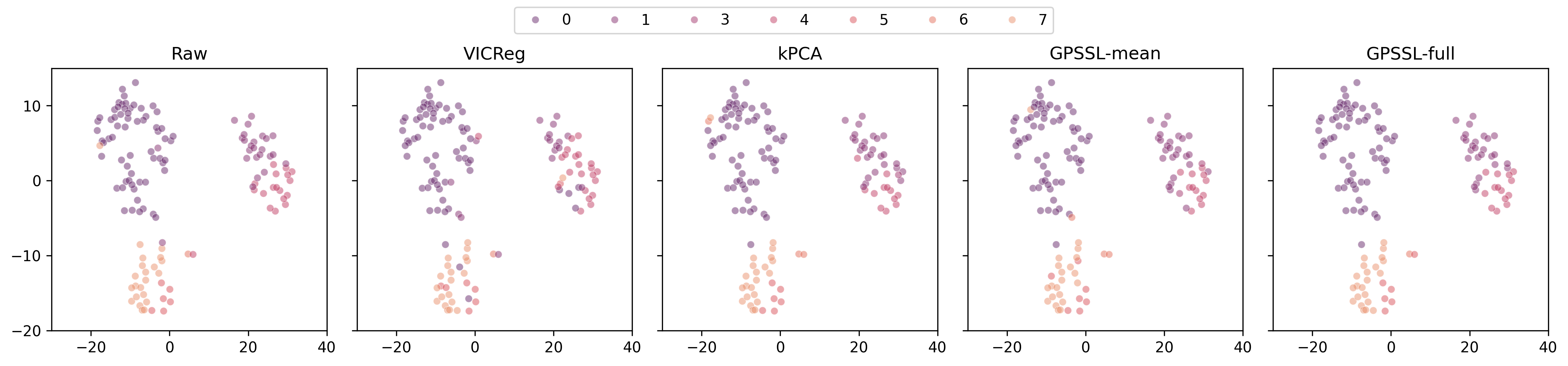}
    \caption{Classification results on ecoli data, comparing original data, kernel PCA, VICReg, GPSSL-mean, and GPSSL-full.}
    \label{fig:ecoli_comparison}
\end{figure}

\begin{figure}[h]
    \centering
    \includegraphics*[width=0.45\textwidth]{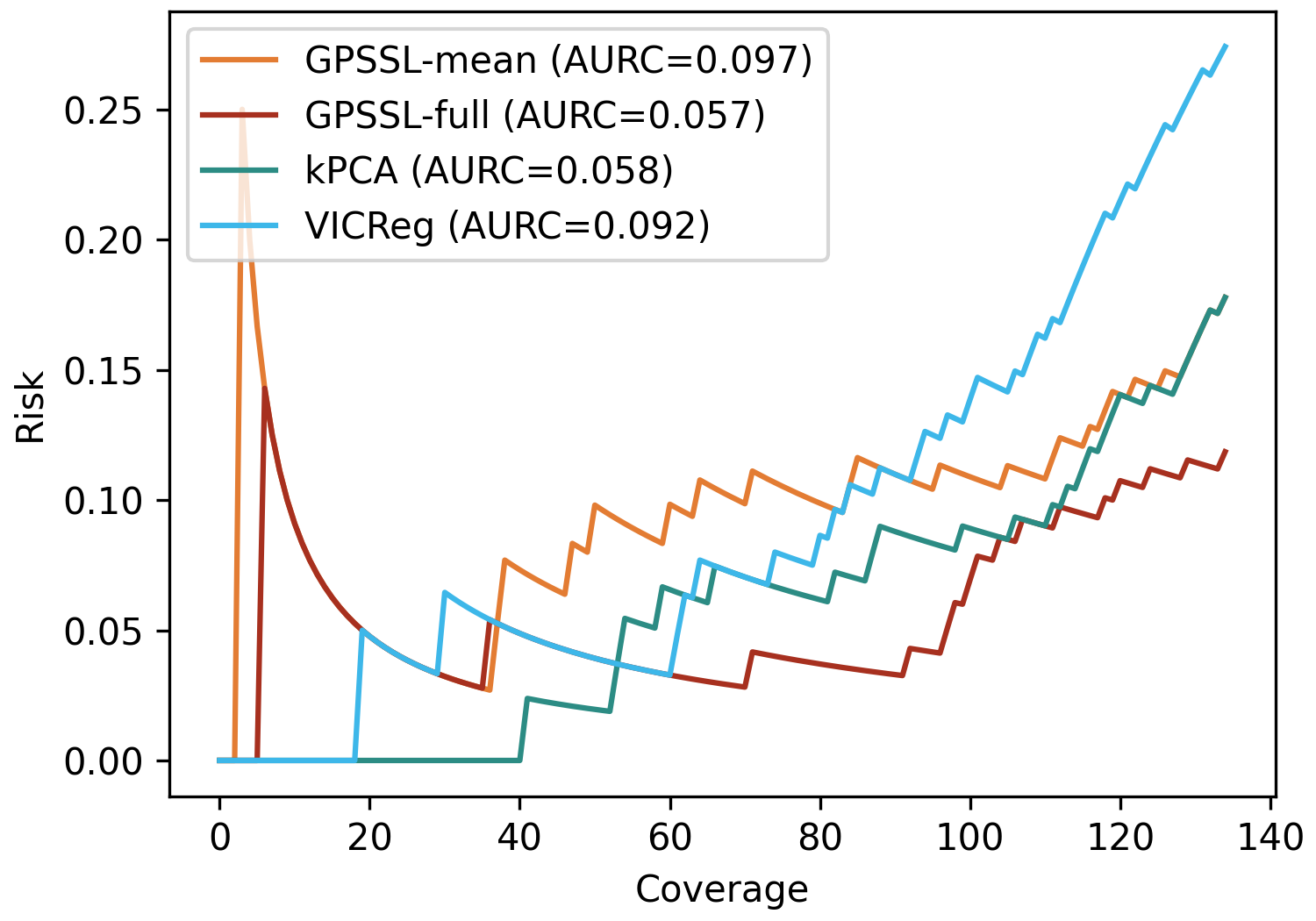}
    \caption{Risk coverage curve of GPSSL and the benchmark methods on ecoli data.}
    \label{fig:ecoli_rc}
\end{figure}

\clearpage \newpage
\begin{figure}[H]
    \centering
    \includegraphics*[width=0.8\textwidth]{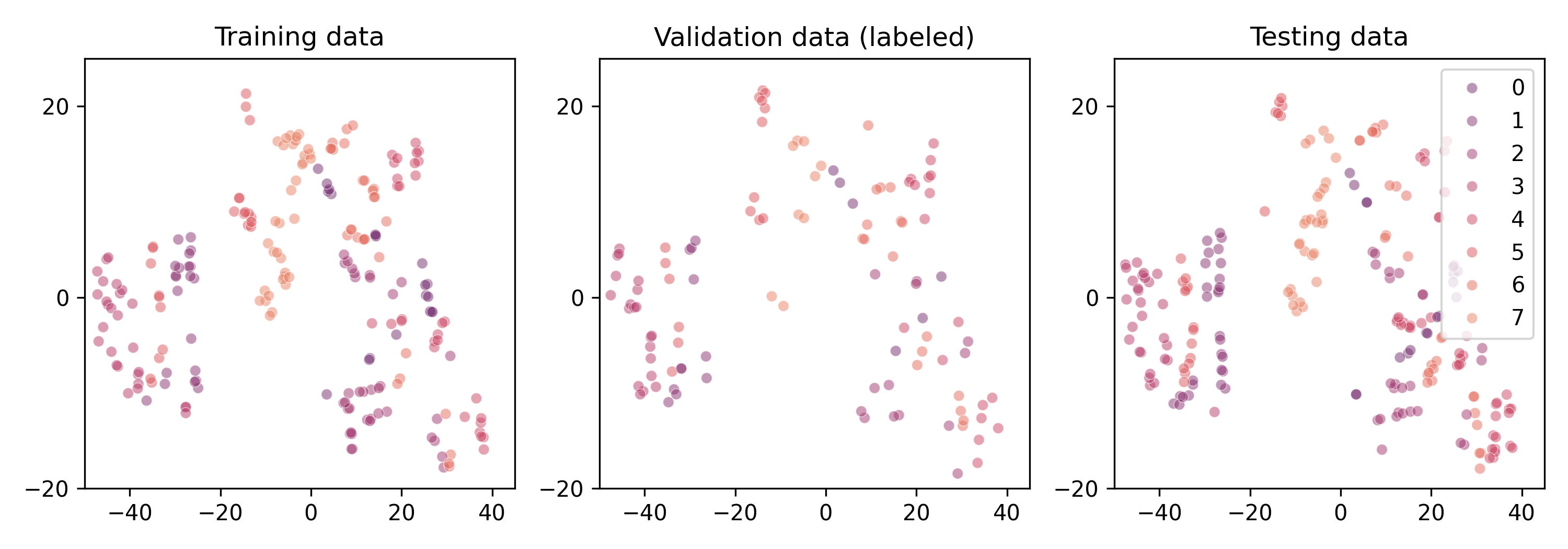}
    \caption{Mice protein data.}
    \label{fig:mice_data}
\end{figure}

\begin{figure}[h]
    \centering
    \includegraphics*[width=\textwidth]{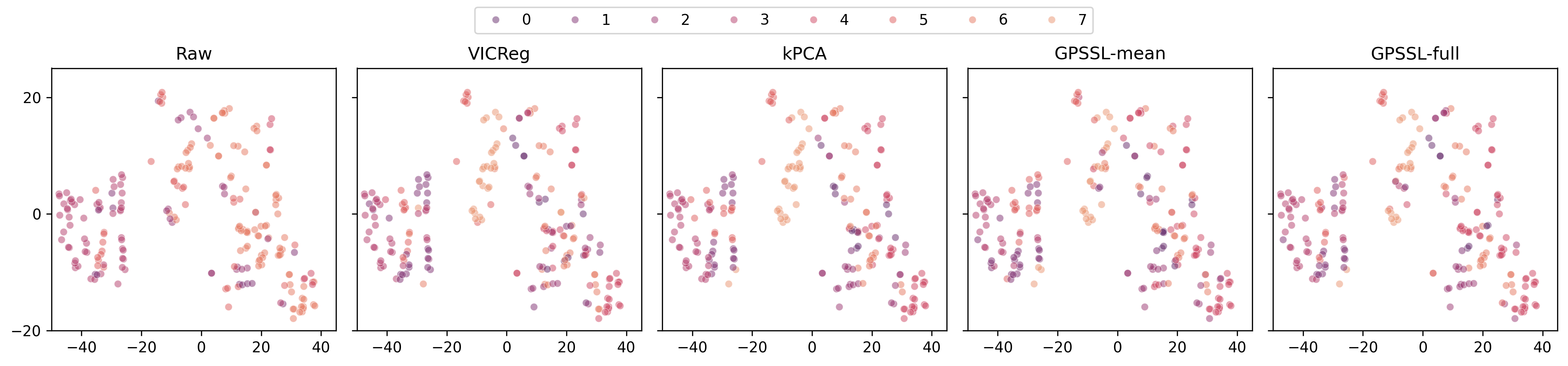}
    \caption{Classification results on mice protein data, comparing original data, kernel PCA, VICReg, GPSSL-mean, and GPSSL-full.}
    \label{fig:mice_comparison}
\end{figure}

\begin{figure}[h]
    \centering
    \includegraphics*[width=0.45\textwidth]{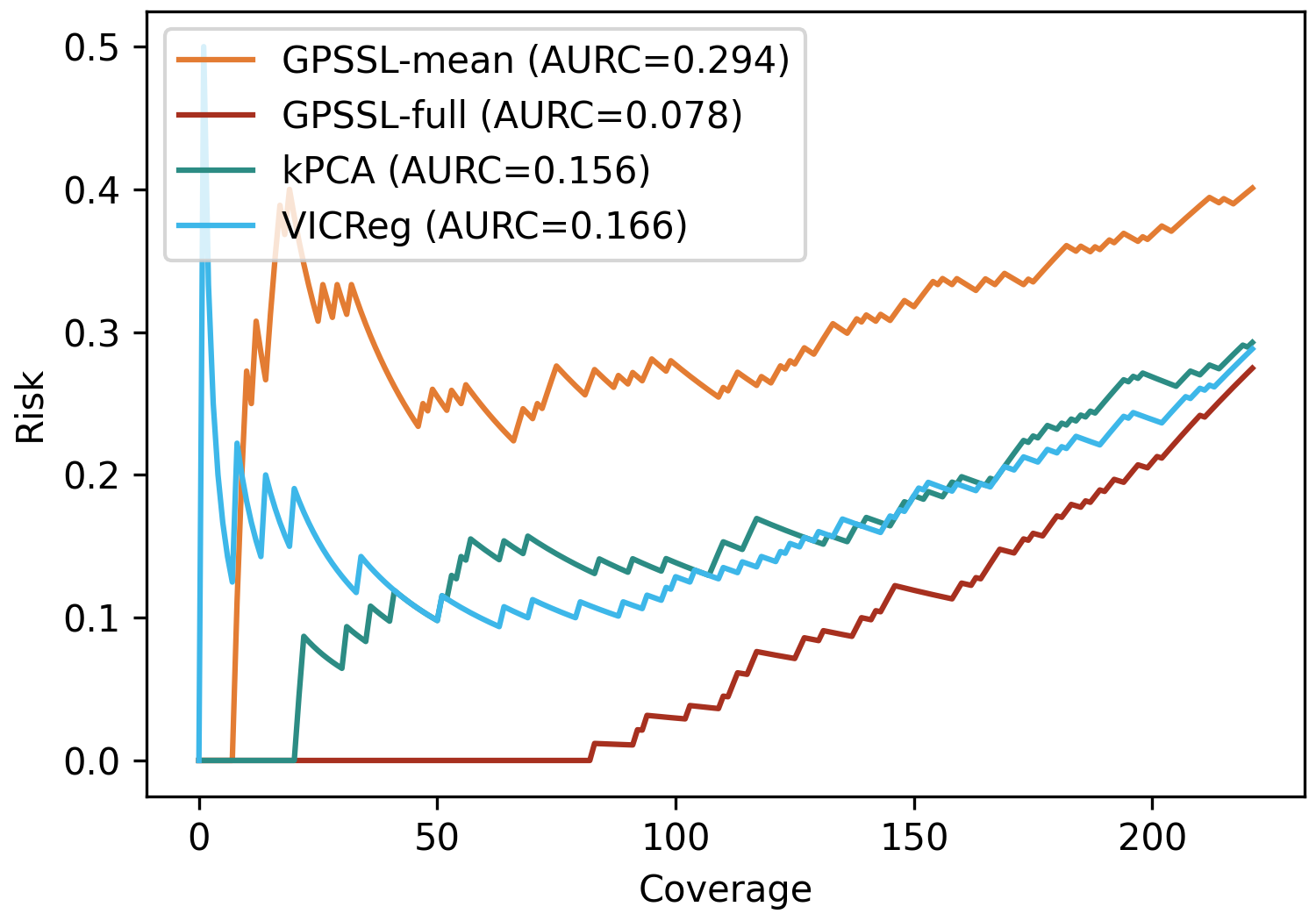}
    \caption{Risk coverage curve of GPSSL and the benchmark methods on mice protein data.}
    \label{fig:mice_rc}
\end{figure}



\end{document}